\documentclass[journal]{IEEEtran}

\usepackage{amsmath}
\usepackage{amsfonts}
\usepackage{amssymb}

\usepackage{amsthm}
\usepackage[ruled,vlined]{algorithm2e}
\usepackage{algorithmic}
\usepackage{mathtools}
\usepackage{tabularx}
\usepackage{graphicx}
\usepackage{subfigure}
\usepackage{enumerate}
\usepackage{float}
\usepackage{url}
\usepackage{verbatim}
\usepackage{mathrsfs}  
\usepackage{cite}
\usepackage[linkcolor=black,citecolor=black,urlcolor=black,colorlinks=true]{hyperref}
\usepackage{booktabs}
\usepackage{multirow}
\usepackage{graphicx}
\usepackage[skip=3pt, font={small}]{caption}

\graphicspath{{figures/}}
\IEEEoverridecommandlockouts

\DeclareMathOperator*{\argmax}{arg\,max}

\author{Jialiang Hou*, Xin Zhou*, Neng Pan, Ang Li, Yuxiang Guan, Chao Xu, \\ Zhongxue Gan$^{\dagger}$, and Fei Gao$^{\dagger}$% <-this % stops a space
	\thanks{* Indicates equal contribution. \par $\dagger$ Corresponding author: Fei Gao, Zhongxue Gan.}%
	\thanks{This work was supported by the National Natural Science Foundation of China under Grant 62322314, Shanghai Municipal Science and Technology Major Project 2021SHZDZX0103.}
	\thanks{Jialiang Hou is with the Academy for Engineering and Technology, Fudan University, Shanghai 200433, China, and also with the Huzhou Institute of Zhejiang University, Huzhou 313000, China (e-mail: jlhou19@fudan.edu.cn). }%
	\thanks{Fei Gao, Xin Zhou, Neng Pan, and Chao Xu are with the Institute of Cyber-Systems and Control, Zhejiang University, Hangzhou 310027, China, and also with the Huzhou Institute of Zhejiang University, Huzhou 313000, China (e-mail: fgaoaa@zju.edu.cn).}%
	\thanks{Zhongxue Gan and Yuxiang Guan are with the Academy for Engineering and Technology, Fudan University, Shanghai 200433, China (e-mail:  ganzhongxue@fudan.edu.cn).}%
	\thanks{Ang Li is with the School of Aeronautic Science and Engineering, Beihang University, Beijing 100191, China, and also with the Huzhou Institute of Zhejiang University, Huzhou 313000, China.}%
}

\title{\LARGE \bf
	Primitive-Swarm: An Ultra-lightweight and Scalable Planner for Large-scale Aerial Swarms
}

\makeatletter
\let\@oldmaketitle\@maketitle% Store \@maketitle
\renewcommand{\@maketitle}{\@oldmaketitle
	\vspace{0.3cm}
	\centering
	\setcounter{figure}{0}
	\begin{minipage}{0.9\linewidth}
		\includegraphics[width=1.0\textwidth]{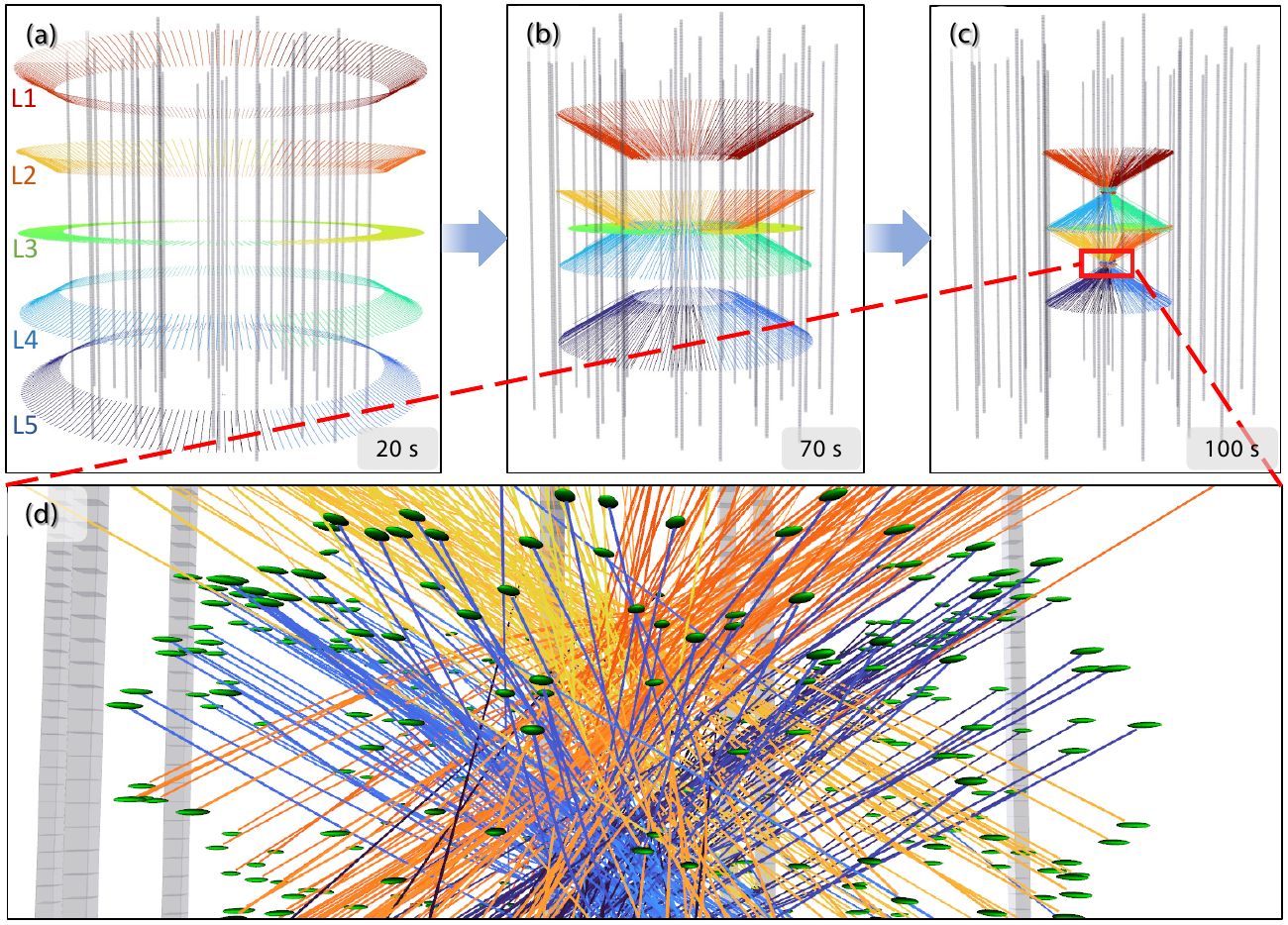}
		\vspace{-0.2cm}
		\captionof{figure}{\label{pic:E1000}Large-scale (\textbf{1000 drones}) air traffic simulation in an unknown environment. The green ellipsoids represent the drones, while the gray pillars represent skyscrapers serving as obstacles. The colored curves illustrate the executed trajectories of the drones. Drones start from five layers L1 to L5, as shown in Figure (a), and then fly to layers L4, L5, L3, L1, L2, respectively. It is notable that all drones operate independently in separate threads, following a decentralized and asynchronous architecture. }
	\end{minipage}
	\vspace{-0.8cm}
}

\makeatother

\begin{document}

    \maketitle

\begin{abstract}

Achieving large-scale aerial swarms is challenging due to the inherent contradictions in balancing computational efficiency and scalability. 
This paper introduces \textit{Primitive-Swarm}, an ultra-lightweight and scalable planner designed specifically for large-scale autonomous aerial swarms.
The proposed approach adopts a decentralized and asynchronous replanning strategy.
Within it is a novel motion primitive library consisting of time-optimal and dynamically feasible trajectories.
They are generated utlizing a novel time-optimial path parameterization algorithm based on reachability analysis (TOPP-RA).
Then, a rapid collision checking mechanism is developed by associating the motion primitives with the discrete surrounding space according to conflicts.
By considering both spatial and temporal conflicts, the mechanism handles robot-obstacle and robot-robot collisions simultaneously. Then, during a replanning process, each robot selects the safe and minimum cost trajectory from the library based on user-defined requirements. Both the time-optimal motion primitive library and the occupancy information are computed offline, turning a time-consuming optimization problem into a linear-complexity selection problem. 
This enables the planner to comprehensively explore the non-convex, discontinuous 3-D safe space filled with numerous obstacles and robots, effectively identifying the best hidden path. Benchmark comparisons demonstrate that our method achieves the shortest flight time and traveled distance with a computation time of less than 1 ms in dense environments.
Super large-scale swarm simulations, involving up to 1000 robots, running in real-time, verify the scalability of our method. Real-world experiments validate the feasibility and robustness of our approach. The code will be released to foster community collaboration.

\end{abstract}

%\textbf{\textit{Index Terms}-- Aerial Swarm}

%%%%%%%%%%%%%%%%%%%%%%%%%%%%%%%%%%%%%%%%%%%%%%%%%%%%%%%%%%%%%%%%%%%%%%%%%%%%%%%%

\section{INTRODUCTION}

Large-scale aerial swarms offer immense potential for addressing various challenging tasks, including air traffic and space colonization \cite{yang2018grand}. To achieve these objectives, a fundamental requirement is the creation of a planner that can generate trajectories for each robot, ensuring safe and reliable navigation towards their individual targets.

In the context of swarm planning, each robot must navigate not only to avoid other members within the swarm but also to effectively avoid collisions with obstacles present in the environment. As the number of robots and the density of obstacles increase, the interactions among robots within the continuously expanding joint state space, combined with frequent robot-obstacle interactions, lead to a combinatorial explosion, rendering real-time online trajectory generation practically infeasible. Despite the significant efforts devoted by researchers in this area, the absence of an ultra-lightweight and scalable planner tailored for large-scale autonomous aerial swarms remains evident. To address this research gap, we propose a swarm planner called \textit{Primitive-Swarm}, which aims to reduce robot-robot and robot-obstacle interactions to formulate low-complexity trajectory generation. 

Swarm planning algorithms face inherent contradictions in computational efficiency and scalability due to robot interactions. To develop an ideal swarm planner, reducing these interactions is essential, where the key is breaking down the high-dimensional trajectory generation problem into easier-to-solve local sub-problems. For this purpose, this work employs motion primitive-based methods in a decentralized and asynchronous manner, stitching together simple motion primitives to form a complex global trajectory. This transformation reduces the high-dimensional online trajectory planning problem to several linear-complexity selection problems while avoiding high-frequency robot-obstacle interactions in problem-solving.

In practise, typical primitive-based methods generate the primitive library mainly by uniformly sampling either in the state or control spaces of a robot (see Section II-B for an extensive literature review). Additionally, collision checking is typically performed individually for each primitive via sample points. 
This strategy requires map inflation or extra data structure like kd-trees to account for robot size, and requires explicit memory storage for each sensor frame, as each gird or point of the surrounding objects may be indexed repeatedly by different primitives. 
As the number of primitives increases, the time complexity of these methods grows.
It becomes a limitation for resource-limited platforms, which may only allow a small set of primitives for real-time usage, consequently compromising the possibility of finding feasible candidate trajectories. 
Moreover, most of these methods only support fixed time allocation for trajectories, leading to a lack of time optimality, which limits the quality of handling both space and time conflicts among a team of moving robots.

Under the aforementioned constraints, the development of a primitive-based planning framework specifically tailored for swarm robots becomes important. This swarm planning framework must fulfill the following requirements: 1) The primitive generation module should be capable of producing high-optimality and dynamically feasible trajectories while maintaining ultra-lightweight online computational costs. 2) The collision checking module must efficiently handle robot-obstacle and robot-robot conflicts, ensuring low computational costs for both environment representation and primitive selection, and then generates high-quality candidate trajectories. 3) The trajectory should accommodate user-defined requirements across various aspects. To address these requirements, we have designed the corresponding modules as follows:

1) Primitive generation module: We construct an innovative motion primitive library offline using time-optimal path parameterization (TOPP) to reduce the computational costs for online planning. This library comprises various primitives that consider time optimality and the robot's dynamical constraints.

2) Collision checking module: For efficient collision checking, we devise an index system which stores the occupancy status between primitives and the surrounding space, referred to as \textit{spatial and spatio-temporal occupancy relationships}. 
% It is used for selecting unsafe trajectories within the primitive library concerning obstacles or other drones' trajectories, respectively. 
In contrast to the traditional trajectory-to-environment approaches that index and check collisions in the environment from sampled points on trajectories, our method takes an environment-to-trajectory approach, which indexes and labels unsafe trajectories from environmental information based on the occupancy relationships. 
Note that in this paper the word \textit{environment} includes obstacles and other robots. 
Robot-obstacle and robot-robot collision avoidance are achieved by leveraging spatial and spatio-temporal occupancy relationships, respectively.
Furthermore, by constructing these occupancy relationships offline, we achieve remarkably high online search efficiency. 
This makes our computational cost dependent only on the number of point clouds and other robots' trajectories, not on the number of primitives, allowing us to set a sufficiently high number of primitives to ensure high-quality candidates, even on resource-limited onboard processors. 
Furthermore, our collision checking method operates directly on raw point clouds without the need for computationally expensive kd-tree construction or obstacle inflation. 

3) Trajectory selection module: Our design of a trajectory selection module incorporates goal-approaching progress and boundary constraints, enabling each robot to approach its global goal as swiftly as possible while respecting the environment's boundaries. Importantly, the collision checking and trajectory selection processes are entirely decoupled in our planning framework. The complexity of trajectory selection remains unaffected by collision checking, thus enhancing the scalability of the swarm.

In summary, all robots share the same motion primitive library based on which the collision checking module efficiently handles robot-obstacle and robot-robot conflicts, marking unsafe primitives. During each replanning step, each robot selects the minimum cost trajectory from safe primitives for execution. This planner minimizes unnecessary online computational costs and demonstrates the capability to be deployed on large-scale swarms.

We compare our method with multiple representative state-of-the-art planners \cite{park2020efficient, tordesillas2021mader, zhou2021ego, zhou2022swarm, kondo2023robust, liu2018towards, adajania2023amswarmx}. The results demonstrate that our method can not only generate trajectories with the shortest flight time and distance but also require the lowest computational cost in dense environments. We implement super large-scale autonomous swarms up to $1000$ robots as depicted in Fig.~\ref{pic:E1000}, which validates the scalability of our mehtod. Furthermore, we validate the feasibility and robustness of our method on SWaP (Size, Weight, and Power) constrained quadrotor platforms in the real world. The contributions of this paper are summarized as follows:

\begin{enumerate}
	\item A novel motion primitive library that generates time-optimal and dynamically feasible trajectories offline, leading to reduced online computational costs.
        \item A fast collision checking method that pre-computes spatial and spatio-temporal occupancy relationships to handle both robot-obstacle and robot-robot collisions in batches.
	\item An ultra-lightweight and scalable planner that transforms high-dimensional trajectory generation into several linear-complexity selection problems.
	\item Open-source code\footnote{https://github.com/ZJU-FAST-Lab/Primitive-Planner} of our system that is extensively validated by simulations and real-world experiments. To the best of our knowledge, our work is the first to handle a swarm of up to 1000 agents while considering high-order dynamics and unknown environments, even achieved the shortest flight time.
\end{enumerate}

\section{RELATED WORK}
\label{sec:related_work}

\subsection{Trajectory Planning for Aerial Swarms}
\label{sec:RW_trajectory_planning}

Bio-inspired flocking algorithms \cite{reynolds1987flocks, vasarhelyi2018optimized} plan trajectories for groups of robots based on simple rules, but they face challenges in dense environments and precise individual robot control. 
Velocity obstacle (VO)-based methods \cite{fiorini1998motion, van2008reciprocal, van2011reciprocal} compute feasible velocities for robots to avoid collisions, but their first-order integrator dynamics models and constant velocity assumptions are not suitable for highly agile quadrotors.
In order to achieve rapid swarm motions in dense real-world environments, some methods \cite{augugliaro2012generation, mellinger2012mixed, kushleyev2013towards, honig2018trajectory, park2020efficient} adopt a centralized approach to jointly plan the global trajectories of all robots. Augugliaro and Mellinger \cite{augugliaro2012generation,mellinger2012mixed, kushleyev2013towards} formulate joint planning as a Sequential Convex Program (SCP) and a Mixed Integer Quadratic Program (MIQP), respectively. H{\"o}nig and Park \cite{honig2018trajectory, park2020efficient} use $\rm{B\acute{e}zier}$ curves and Bernstein polynomials to generate safe and conservative trajectories, respectively. However, these methods rely on known environments, global localization devices, and entail high computational costs, thereby limiting the full-autonomy and scalability of swarm systems.

In contrast, some methods \cite{chen2015decoupled, luis2019trajectory, liu2018towards} adopt a decentralized approach to improve computational efficiency. However, the assumption of synchronous replanning limits their applicability in real-world scenarios. Tordesillas et al. \cite{tordesillas2021mader} propose a decentralized and asynchronous algorithm that handles multi-robot and dynamic obstacles in convex obstacle environments.
Zhou et al. \cite{zhou2021ego} establish a fully autonomous decentralized quadrotor swarm system suitable for non-convex unknown environments. However, the challenge of temporal optimization leads to twisted trajectories when multiple robots encounter each other. To address this issue, Zhou et al. \cite{zhou2022swarm} deploy a spatial-temporal trajectory optimization algorithm \cite{wang2022geometrically}.

\begin{figure}[t]
	\centering
	\includegraphics[width=1.0\linewidth]{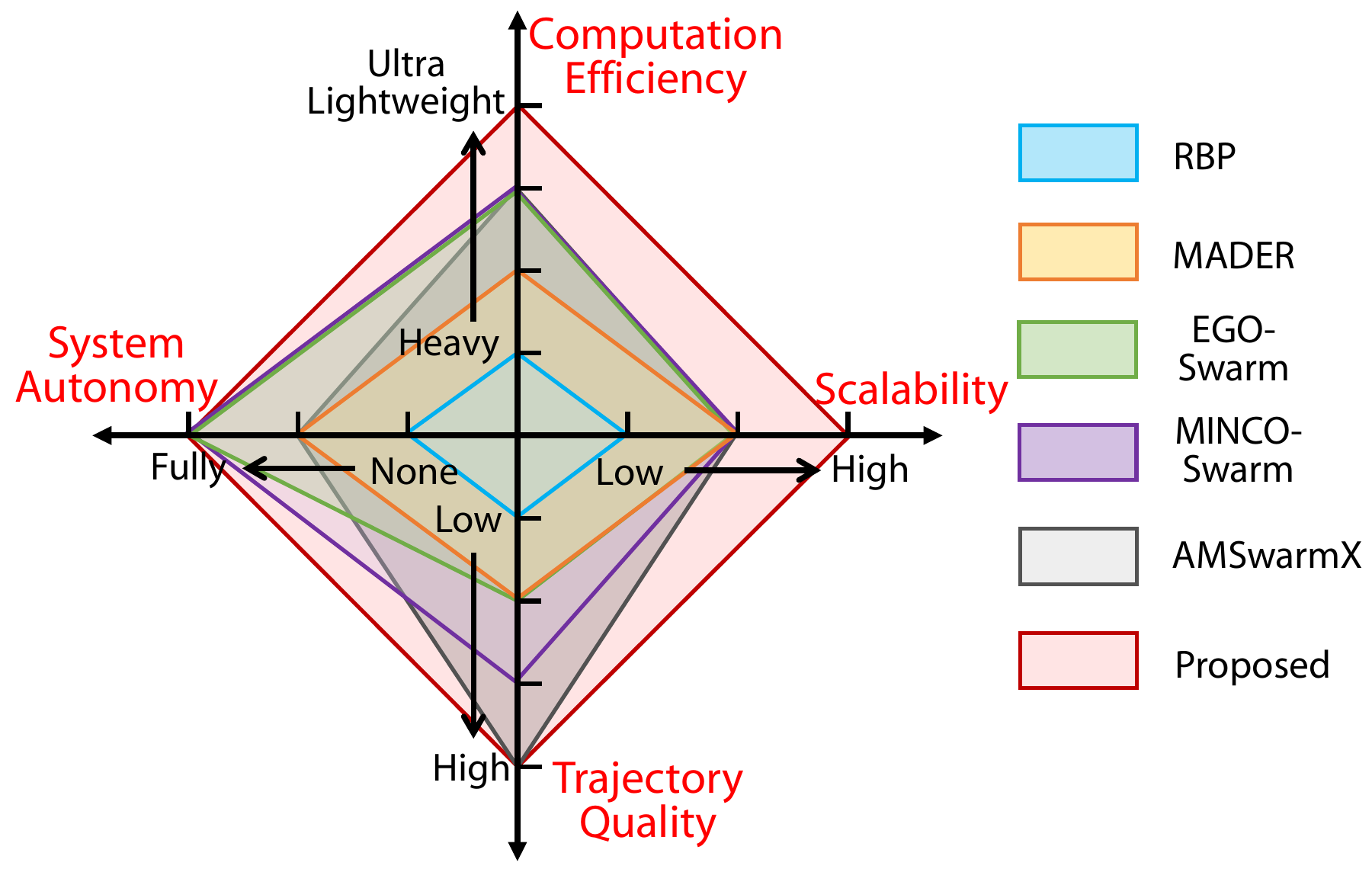}
	\caption{Qualitative comparison with RBP\cite{park2020efficient}, MADER\cite{tordesillas2021mader}, EGO-Swarm\cite{zhou2021ego}, MINCO-Swarm\cite{zhou2022swarm}, and AMSwarmX\cite{adajania2023amswarmx}. Axis ticks from inside to outside represent: Computation efficiency - heavyweight, moderate, lightweight, ultra lightweight; Scalability - low, medium, high; Trajectory quality - low, medium, high, very high; System Autonomy - none, partially autonomous, fully autonomous. Please refer to Section \ref{sec:RW_trajectory_planning} for the detailed discussion of the qualitative comparison.}
	\label{pic:swarm_comparison}
	\vspace{-0.5cm}
\end{figure}

In our previous work \cite{hou2022enhanced}, we proposed a group planning strategy aimed at enhancing the scalability of swarm systems.
However, these methods \cite{tordesillas2021mader,zhou2021ego,zhou2022swarm, hou2022enhanced} require complex environment representation, such as inflated grid maps or convex obstacles, and demand good initial trajectories for nonlinear trajectory optimization, leading to substantial onboard computation.
Furthermore, the inclusion of non-convex collision avoidance terms for robot-obstacle and robot-robot interactions makes the trajectory susceptible to infeasible local minima, resulting in replanning failure. Additionally, the generated trajectory is searched only within a limited portion of the solution space around the initial values.

The proposed method effectively mitigates these challenges and introduces an ultra-lightweight algorithm suitable for large-scale autonomous swarms. Comprehensive benchmark comparisons with representative state-of-the-art methods are conducted in Section \ref{sec:evaluation}, which are then summarized in Fig. \ref{pic:swarm_comparison} to give an intuitive knowledge of some main characteristics of the proposed method. In Fig. \ref{pic:swarm_comparison}, \textit{Computation Efficiency} is evaluated in Table \ref{tab:comparison_empty}, Fig. \ref{pic:obs_collision_time}, \ref{pic:agent_collision_time} and \ref{pic:30_radnom_comp}c. \textit{Trajectory Quality} is compared in Fig. \ref{pic:comparison_single} to \ref{pic:30_radnom_comp} where the proposed method generates the smoothest trajectory while maintaining the maximum speed along the whole flight. Flight time and flight distance in Table \ref{tab:comparison_empty} are also metrics of trajectory quality. High \textit{Scalability} is the main characteristic of this work as featured in Fig. \ref{pic:E1000} and \ref{pic:1000_circle} and analysed in Section \ref{sec:large_scale}. RBP receives a relatively low grade in this metric for its high computation time (Table \ref{tab:comparison_empty}). \textit{System Autonomy} refers to the ability of navigating in the real world relying on only onboard sensing and computation. RBP is a centralized trajectory planner that requires a high-performance ground computer and a pre-built map, thus it lacks autonomy. MADER and AMSwarmX are decentralized swarm planners validated with pre-built maps in simulation only. Although both of them have the potential of using online generated maps, a lot of future work is required. EGO-Swarm, MINCO-Swarm and the proposed method are all fully autonomous methods validated in real-world experiments.

\subsection{Motion Primitive-based Trajectory Planning}

% \usepackage{multirow}
% \usepackage{graphicx}
\begin{table}[]
	\centering
	\caption{Evaluation of Motion Primitive-based Methods}
    \label{tab:primitive}
    \renewcommand{\arraystretch}{1.3}
    \setlength\tabcolsep{2pt}
	\resizebox{\columnwidth}{!}{%
		\begin{tabular}{|cc|c|c|c|c|c|}
			\hline
			\multicolumn{2}{|c|}{\textbf{Method}} &
			\textbf{\begin{tabular}[c]{@{}c@{}}Primitive \\ Generation\end{tabular}} &
			\textbf{\begin{tabular}[c]{@{}c@{}}Time \\ Parameterized\end{tabular}} &
			\textbf{\begin{tabular}[c]{@{}c@{}}Dynamical\\ Feasiblity\end{tabular}} &
			\textbf{\begin{tabular}[c]{@{}c@{}}Time\\ Optimality\end{tabular}} &
			\textbf{\begin{tabular}[c]{@{}c@{}}Collision\\ Checking\end{tabular}} \\ \hline
			\multicolumn{1}{|c|}{\multirow{3}{*}{\begin{tabular}[c]{@{}c@{}}State\\  Space\end{tabular}}} &
			\cite{zhang2020falco} &
			Offline &
			No &
			\textbackslash{} &
			\multirow{4}{*}{No} &
			Medium \\ \cline{2-5} \cline{7-7} 
			\multicolumn{1}{|c|}{} &
			\cite{ryll2019efficient} &
			\multirow{3}{*}{Online} &
			\multirow{4}{*}{\textbf{Yes}} &
			\multirow{2}{*}{Post-check} &
			&
			\multirow{3}{*}{Slow} \\ \cline{2-2}
			\multicolumn{1}{|c|}{} &
			\cite{yang2021intention,collins2020efficient} &
			&
			&
			&
			&
			\\ \cline{1-2} \cline{5-5}
			\multicolumn{2}{|c|}{Control Space\cite{florence2020integrated}} &
			&
			&
			\multirow{2}{*}{\textbf{\begin{tabular}[c]{@{}c@{}}Satisfied in \\ Generation\end{tabular}}} &
			&
			\\ \cline{1-3} \cline{6-7} 
			\multicolumn{2}{|c|}{\textbf{Proposed}} &
			Offline &
			&
			&
			\textbf{Yes} &
			\textbf{Fast} \\ \hline
		\end{tabular}%
	}
	\vspace{-0.4cm}
\end{table}

\begin{figure*}[t]
	\centering
	\includegraphics[width=1.0\linewidth]{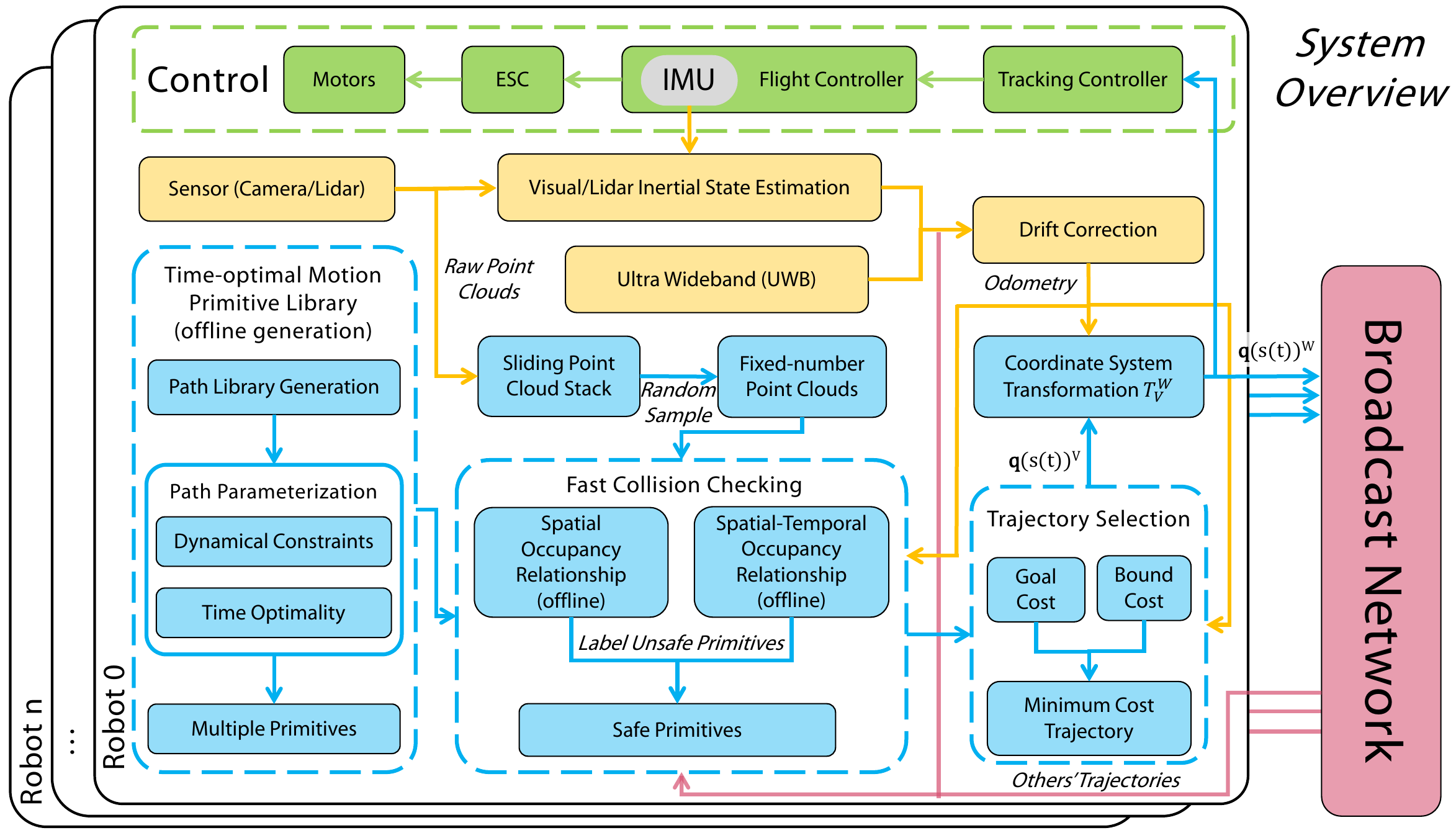}
	\caption{The overview of our decentralized and asynchronous autonomous aerial swarm system, which includes state estimation (yellow), planning (blue), control (green), and communication (red) modules. $\mathcal{W}$ refers to the world coordinate system and $\mathcal{V}$ is called the velocity-aligned frame defined by current drone position and velocity in Sec. \ref{sec:vcs}.}
	\label{pic:system_overview}
	\vspace{-0.2cm}
\end{figure*}

Motion primitive-based methods are commonly employed for generating multiple paths or trajectories for single-robot autonomous navigation. They effectively reduce the planning problem's complexity and can cover a wide solution space simultaneously. 
A summary of typical methods is presented in TABLE \ref{tab:primitive}. The index ``Time Parameterized" means whether the primitive library consists of time-parameterized continuous trajectories, whose higher derivatives (i.e., velocity, acceleration, jerk, and more) can be easily obtained. The index ``Dynamical Feasibility'' records how dynamical feasibility are guaranteed. The index ``Collision Checking'' contains 
the theoretical computation time for both robot-obstacle and robot-robot collision avoidance. If a method being compared is initially designed for robot-obstacle avoidance only, we assume the space swept by nearby drones as a normal static obstacle, thus compatible with the compared method.

Zhang et al. \cite{zhang2020falco} propose an offline motion primitive library that samples fixed-length paths with only position information, lacking dynamical details and not fully exploiting the robot's maneuverability. Collision checking is realized by building an adjacency list that records the occupancy status between primitives and the surrounding space offline during primitive library generation. We draw inspiration in collision checking from \cite{zhang2020falco}.
Ryll et al. \cite{ryll2019efficient} generate multiple fixed-duration min-jerk trajectories by sampling different local end positions along with a start state (position, velocity, and acceleration).
Yang et al. \cite{yang2021intention} perform online sampling of different local end velocities in an action space $a={v_x, v_z, \omega}$, combining them with a start state to generate multiple fixed-duration 8-th order polynomial trajectories, which are subsequently integrated to obtain position information.
These methods \cite{ryll2019efficient, yang2021intention} generate online primitives that include a significant number of dynamically infeasible trajectories, necessitating a rechecking mechanism \cite{mueller2015computationally} to filter out feasible ones.
Collins et al. \cite{collins2020efficient} employ adaptive end velocity sampling based on a robot's reference velocity, enhancing the dynamical reliability of motion primitive libraries.
Florence et al. \cite{florence2020integrated} perform online sampling of a set of constant control variables considering a quadrotor dynamics model. These variables are then integrated forward for a fixed duration to generate dynamically feasible primitives.
The aforementioned methods \cite{ryll2019efficient, yang2021intention, collins2020efficient, florence2020integrated} suffer from the drawback of checking collisions for each primitive individually on a fusion map or kd-tree. As the number of primitives increases, these methods become progressively time-consuming.
In contrast, Bucki et al. \cite{bucki2020rectangular} accelerate collision checking using a pyramid partitioning method, but it results in more conservative trajectories and performs poorly in dense environments.

In summary, the aforementioned methods fail to simultaneously meet three critical requirements: (1) ultra-lightweight online computational cost; (2) time-optimal and dynamically feasible primitives; (3) fast collision checking methods capable of handling both robot-obstacle and robot-robot conflicts in batches. These limitations have hindered the advancement of motion primitive libraries in autonomous navigation. The proposed method is the first to successfully integrate all three aspects and deploy motion primitives on autonomous swarms, to the best of our knowledge.

\subsection{Time-optimal Trajectory Generation }

Time-optimal trajectory generation address the problem of finding the fastest way to traverse a region while adhering to dynamical constraints.
Mellinger et al. \cite{mellinger2011minimum} find the time-optimal trajectory by optimizing time allocation through backtracking gradient descent. Then dynamical feasibility is achieved by temporal scaling. However, both strategies consume significant computation as the number of trajectory pieces grows.
Liu et al. \cite{liu2017planning} managed to achieve comparable performance with only a single scaling operation, but it is only valid for rest-to-rest trajectories.
Sun et al. \cite{sun2021fast} adopt a dual-level optimization scheme by analytically estimating the projected gradient by leveraging the dual solution of the low-level Quadratic Programming (QP), resulting in enhanced precision. However, computing the projected gradient remains less effective.
Richter et al. \cite{bry2015aggressive} address this challenge by treating each duration as an independent variable and employing total duration as a regularization term. They optimize time allocation through gradient descent and ensuring that actuator constraints are met through scaling. However, the optimal time allocation is sensitive to constraints, resulting in the potential to disrupt a trajectory.
Burri et al. \cite{burri2015real} relieve the sensitivity by converting constraints as weighted objectives and optimizing through nonlinear programming, but at the cost of being dependent of initial values and the risk of constraints violation.

\section{System Overview}
\label{sec:system_overview}

\begin{figure}[t]
	\centering
	\includegraphics[width=1.0\linewidth]{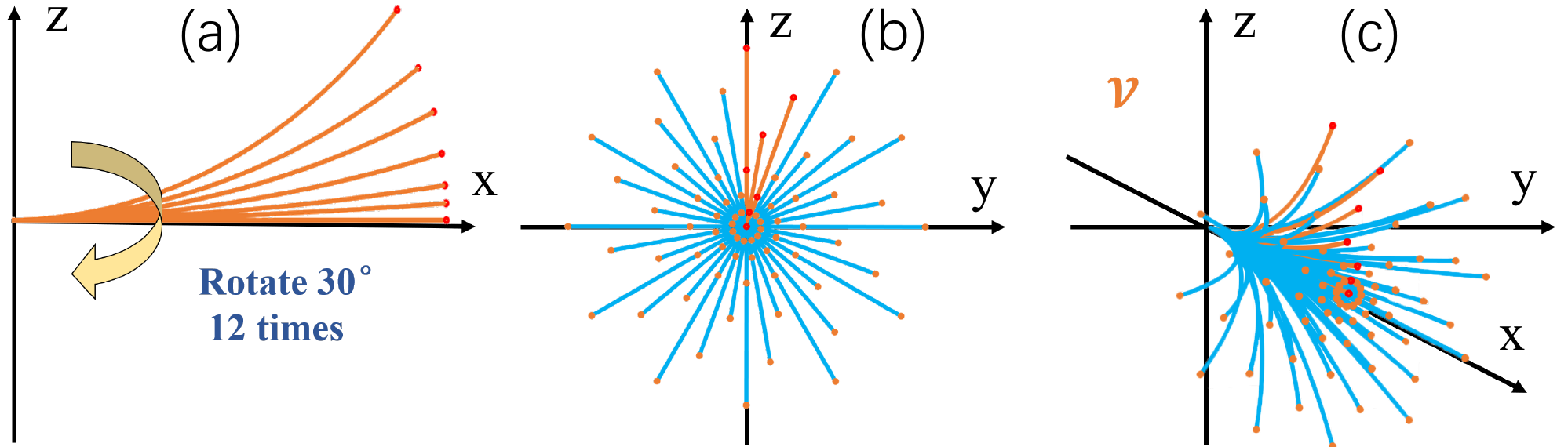}
         \caption{An example of path library generation (Note: \textit{Path} refers to a geometric curve without temporal information). (a) Seven orange arcs representing paths; (b) Front view of the path library obtained by rotating the arcs in the sub-figure (a) around the x-axis with an angle interpolation of $D_{angle}=30^\circ$; (c) Side view of the path library, where both blue and orange arcs represent paths. The origin of all paths coincides with the origin of the velocity-aligned coordinate system $\mathcal{V}$. All paths start tangent to the x-axis. The orange and red dots represent the end points of each path.}
	\label{pic:path_lib}
	\vspace{-0.7cm}
\end{figure}

The proposed decentralized and asynchronous autonomous aerial swarm system is depicted in Fig. \ref{pic:system_overview}. Each robot independently handles state estimation, planning, and control, and communicates its trajectory through a broadcast network, resulting in a loosely coupled system. Detailed implementation information for the state estimation, control, and communication modules is provided in Section \ref{sec:details}.

The planning module receives the output of the state estimation module and generates new trajectories. The planner operates in four steps. First, the motion primitive library module generates multiple time-optimal primitives offline, taking into account dynamical constraints (Section \ref{sec:mpl}).
Second, the sliding point cloud stack stores the raw point clouds of the latest $N_f$ frames. From the stack, fixed-number point clouds are sampled using a random sample method \cite{vitter1984faster} to maintain environment fidelity.
Trajectories of other robots within two \textit{planning horizons} (as defined in Fig. \ref{pic:replan}c) are transmitted to the current robot through the peer-to-peer broadcast network.
The fast collision checking method removes unsafe primitives by querying the spatial and spatio-temporal occupancy relationships (Section \ref{sec:collision_check}).
Then the minimum cost trajectory $\mathbf{q}(t)^\mathcal{V}$ is selected among safe primitives from the velocity frame $\mathcal{V}$ (Section \ref{sec:vcs}) according to user-defined requirements. 
Finally, the trajectory $\mathbf{q}(t)^\mathcal{W}$ in the world coordinate system $\mathcal{W}$ is obtained through the transformation $T_\mathcal{V}^\mathcal{W}$ (Section \ref{sec:re_plan}). The control module then executes the trajectory $\mathbf{q}(t)^\mathcal{W}$.

\section{Time-optimal Motion Primitive Library}
\label{sec:mpl}
We divide the process of generating the time-optimal motion primitive library into two steps: \textit{path library generation} (spatial) and \textit{path parameterization} (temporal). The former involves constructing paths using fundamental geometric primitives. The latter addresses time allocation for the generated paths by solving the Time-Optimal Path Parameterization (TOPP) problem, which aims to find the quickest path traversal while satisfying to dynamic constraints.

\subsection{Path Library Generation}
\label{subsubsec:lib}

The configuration of an $n$-degree-of-freedom robot system is denoted by an $n$-dimensional vector $\mathbf{q} \in \mathbb{R}^n$. According to the principles of differential flatness \cite{mellinger2011minimum}, the configuration space of a quadrotor is represented as follows:
\begin{equation}
\mathbf{q}=[p_x,p_y,p_z,\psi]^T
\end{equation}
% where the $(p_x,p_y,p_z)^T$ is the center of mass in the world coordinate system and $\psi$ is the yaw angle. 
where $(p_x,p_y,p_z)^T$ denotes the center of mass in the global coordinate system and $\psi$ denotes the yaw angle.
In actual flight, $\psi$ can be arbitrarily set \cite{mellinger2011minimum}, or it is usually set to be the angle between the projection of the x-axis of the Body frame on the horizontal plane and the x-axis of the World frame \cite{2017Differential}.

With above insight, we construct a path library $\mathcal{P}(i_{prim}, s): \mathbb{N} \times \mathbb{R}^+ \rightarrow \mathbb{R}^3$ by rotating multiple arcs with the parameters: the number of arcs $N_a$ and rotation angle interpolation $D_{angle}$. These arcs begin at the origin of the coordinate system and are tangent to the x-axis at the origin, as shown in Fig. \ref{pic:path_lib}a.
They share a common length $l$ while having distinct radii $r$.  The initial angle $\theta$ is the rotation of an arc around x-axis, as shown in Fig. \ref{pic:path_lib}b. 
Here, $i_{prim} \in \mathbb{N}$ is the index of a primitive and $s \in \mathbb{R}^+$ is the function parameter, which is used only to retrieve path values and has no actual physical meaning.
In the following content, we use the notation $\mathcal{P}$ that omits the argument to stands for $\mathcal{P}(i_{prim}, s)$.
For instance, in Fig. \ref{pic:path_lib}a, we choose $N_a=7$ arcs, with the following parameters:
\begin{equation}
\begin{gathered}
l = 5~m,~~
r \in \{6, 8, 12, 20, 36, 78, \infty\}~m,\\
\theta \in \{0^\circ, -10^\circ, -20^\circ, 0^\circ, -10^\circ, -20^\circ, 0^\circ\},
\end{gathered}
\end{equation} 
where $r=\infty$ corresponds to a straight line. We rotate these arcs around the x-axis using $D_{angle} = 30^\circ$, resulting in a path library comprising $73$ geometric paths (Fig. \ref{pic:path_lib}b-c).
Varying the rotation start angles $\theta$ enhances the spatial distribution coverage of the path library, ensuring comprehensive coverage of the robot's anticipated travel area. Adjusting parameters like $N_a$, $D_{angle}$, $l$, $r$, and $\theta$ enables the generation of diverse path libraries that align with user-defined needs and environmental considerations.

\subsection{Path Parameterization}

To parameterize a generated geometric path from the path library $\mathbf{q}(s)_{s\in[0, s_{end}]}\in\mathcal{P}$, the time parameterization process involves establishing a scalar function $s(t):[0, T] \rightarrow [0, s_{end}]$, enabling the reconstruction of a trajectory $\mathbf{q}(s(t))_{t\in [0, T]}$.
Moreover, in the following paper, $\square'$ denotes differentiation with respect to the path parameter $s$, and $\dot{\square}$ denotes differentiation with respect to time $t$.
As depicted in Algorithm \ref{alg:mpl}, for each geometry path $\mathbf{q}(s)\in\mathcal{P}$, distinct start and end velocities of the quadrotors are set according to:
\begin{equation}
\label{eq:start_end_speed}
\begin{gathered}
\lVert \dot{\mathbf{q}}(s_0) \rVert \in \{0,~0.1,~...,~v_{max}\}~m/s,\\
\lVert \dot{\mathbf{q}}(s_{end}) \rVert = 0~m/s,
\end{gathered}
\end{equation} 
where the velocity direction aligns with the path tangent. Here, $v_{max}$ represents the permissible maximum robot velocity. Varied start velocities $\dot{\mathbf{q}}(s_0)$ are used for velocity selection (see Section \ref{sec:traj_select}) during each replanning. The negative effect caused by such small velocity discontinuity can be safely neglected through our experiment. The setting $\dot{\mathbf{q}}(s_{end}) = \mathbf{0}~m/s$ is employed to ensure numerical stability.

\begin{figure}[t]
	\centering
	\includegraphics[width=0.9\linewidth]{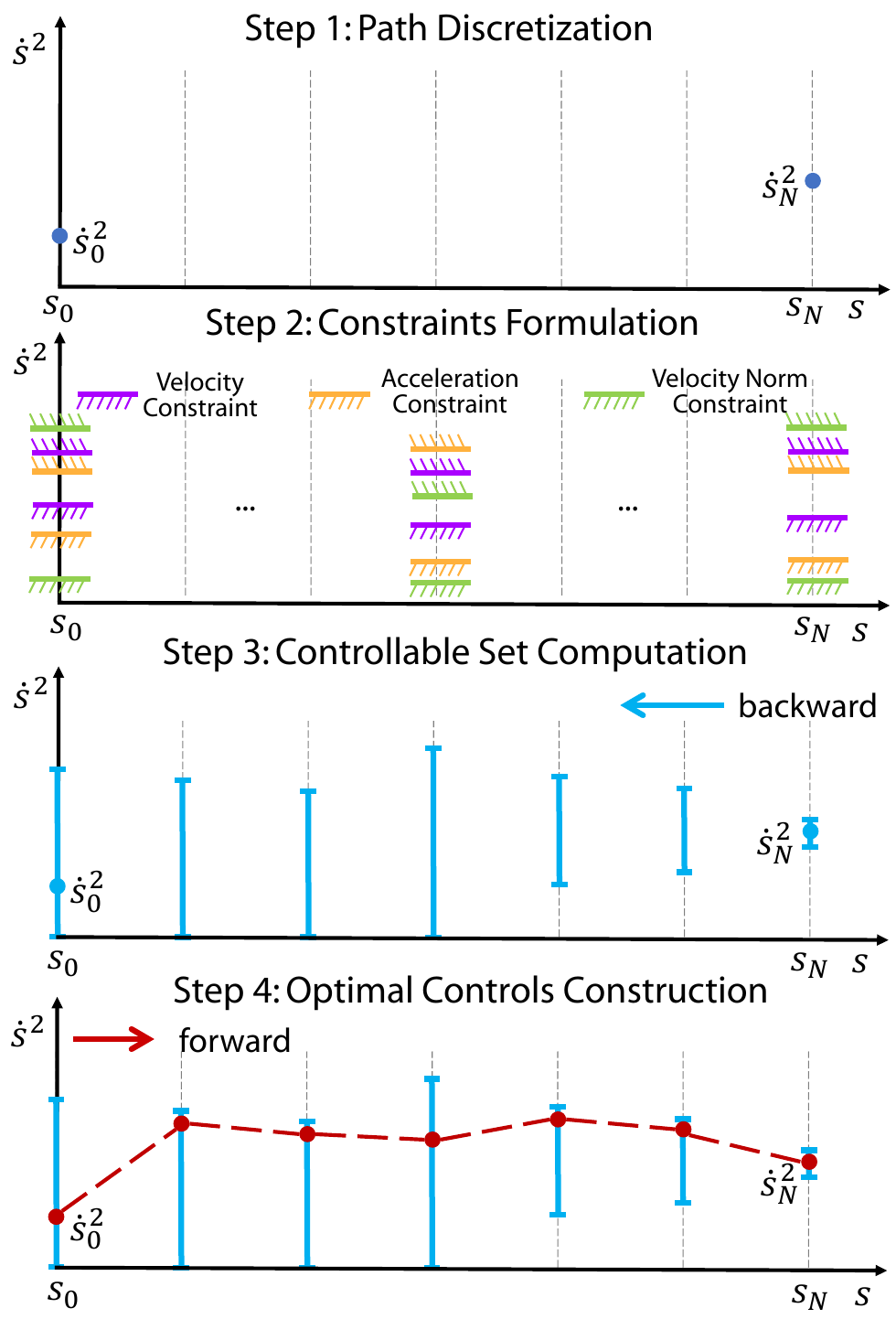}
        \caption{Time parameterization process for path $\mathbf{q}(s)$ with specified start and end velocities, $\dot{s}_0$ and $\dot{s}_N$. Step 1 involves discretizing $s$. In Step 2, dynamical constraints are applied to each discrete position $s_i$. Step 3 encompasses the sequential computation of the controllable set $\mathcal{K}_i(\mathbb{I}_N)$ (illustrated as blue intervals) in a backward manner, starting from the discrete position $s_N$. In Step 4, we iteratively select the largest admissible control $u_i^*$ from the discrete position $s_0$ in a forward manner. The optimal state $x_{i+1}^* := x_i^* + 2\Delta_iu_i^*$ (represented as red points) ensures that it remains within the corresponding controllable set $\mathcal{K}_{i+1}(\mathbb{I}_N)$.}
	\label{pic:path_para}
	\vspace{-0.2cm}
\end{figure}

For a given path $\mathbf{q}(s)$, the initial and final velocities along the path, $\dot{s}_0$ and $\dot{s}_N$, are specified. We employ an effective and robust time-optimal parameterization technique known as TOPP-RA \cite{pham2018new}. The key concept of this approach lies in iteratively calculating reachable and controllable sets at discrete positions $\{s_i\}$ along the path through the solution of compact Linear Programs (LPs).
Illustrated in Fig. \ref{pic:path_para}, TOPP-RA involves four fundamental stages: path discretization, formulation of constraints, computation of controllable sets (backward pass), and construction of optimal controls (forward pass). This methodology offers a comprehensive strategy to implement time parameterization with high efficiency and reliability.

\begin{algorithm}
    \begin{algorithmic}[1]
        \REQUIRE Path Library $\mathcal{P}$
        \ENSURE Time-optimal motion primitive library $\mathcal{P}(t)$
        
        \STATE end speed $\dot{\mathbf{q}}(s_{end}) = \mathbf{0}$
        \FOR{each path $\mathbf{q}(s)\in\mathcal{P}$}
            \FOR{each start speed $\dot{\mathbf{q}}(s_0)$ in Equ. \ref{eq:start_end_speed}}
                \STATE $\dot{s}_0=\lVert\dot{\mathbf{q}}(s_0)\lVert/\lVert\mathbf{q}'(s_0)\lVert$
                \STATE $\dot{s}_{end}=\lVert\dot{\mathbf{q}}(s_{end})\lVert/\lVert\mathbf{q}'(s_{end})\lVert$
                \STATE $\mathbf{q}(s(t))$ = TOPP-RA $(\mathbf{q}(s), \dot{s_0}, \dot{s}_{end})$
                \STATE $\mathcal{P}(t)$.push\_back($\mathbf{q}(s(t))$)		
            \ENDFOR
        \ENDFOR
	
        \RETURN $\mathcal{P}(t)$
    \end{algorithmic}
    \caption{Parameterization for Path Library $\mathcal{P}$}
    \label{alg:mpl}
\end{algorithm}
%\vspace{-0.2cm}

\subsubsection{\textbf{Path Discretization}}
The interval $[0, s_{end}]$ is divided into $N$ segments and $N+1$ grid points
\begin{equation}
0 =: s_0, s_1, ..., s_{N-1}, s_N := s_{end},
\end{equation}
as shown in Fig. \ref{pic:path_para}, Step 1.

Successive differentiation of $\mathbf{q}(s)$ yields the following expressions:
\begin{equation}
	\label{eq:diff}
	\dot{\mathbf{q}}= \mathbf{q}'\dot s, ~\ddot{\mathbf{q}}  = \mathbf{q}''\dot s ^2 + \mathbf{q}'\ddot s.
\end{equation}
Please note the difference between $\square'$ and $\dot{\square}$.

% We define the constant path acceleration $u_i=\ddot{s}_i$ and the squared velocity $x_i=\dot{s}_i^2$ in the interval $[s_i, s_{i+1}]$. They are related as follows
We define the path acceleration $\ddot{s}_i$ and the squared velocity $\dot{s}_i^2$ by
\begin{equation}
u_i := \ddot{s}_i,~x_i := \dot{s}_i^2
\label{eq:uixi}
\end{equation} 
within the interval $[s_i, s_{i+1}]$. 
They are interrelated as follows:

\begin{subequations}
	\label{eq:dxidsi}
	\begin{gather}
	\frac{dx_i}{dt} = \frac{d\dot{s}_i^2}{dt} = 2\dot{s}_i\ddot{s}_i=2\frac{d s_i}{d t}\ddot{s}_i=2\frac{d s_i}{d t}u_i,\\
	d x_i=2 ds_i u_i.
	\end{gather}
\end{subequations}

Equation (\ref{eq:dxidsi}b) establishes a linear connection among $x$, $u$, and $s$:
\begin{equation}
\label{eq:control_state}
{x}_{i+1} = {x}_i + 2 \Delta_i u_i,~~~i=0...N-1,
\end{equation} 
where $\Delta_i:=s_{i+1}-s_i$.

According to (\ref{eq:control_state}), the terms $u_i$ and $x_i$ are referred to as the \textit{control} and \textit{state} of the $i$-th discrete position $s_i$. Any sequence $x_0, u_0$, ..., $x_{N-1}, u_{N-1}, x_{N},u_N$ that satisfies (\ref{eq:control_state}) is denoted as a path parameterization. Our objective is to determine the sequence that minimizes the total time $T$ of the trajectory $\mathbf{q}(s(t))$.

\subsubsection{\textbf{Constraints Formulation}}
The dynamic constraints of the quadrotor are set to
\begin{equation}
\label{eq:sec_constraint}
	|| \dot{\mathbf{q}} || \leq v_{max}, -a_{max} \leq \ddot{ \mathbf{q}}_{\rm \{x,y,z\}} \leq a_{max},
\end{equation}
where $v_{max}$ and $a_{max}$ denote the upper bounds for velocity and acceleration respectively, and $\mathbf{q}_{\rm \{x,y,z\}}$ refers to the individual elements along each axis.
By substituting (\ref{eq:diff}) into (\ref{eq:sec_constraint}), the constraints on $\dot{\mathbf{q}}$ and $\ddot{\mathbf{q}}$ are translated into constraints on $\dot{s}^2$ and $\ddot{s}$ as
\begin{subequations}
	\label{eq:qscons}
	\begin{gather}
	\mathbf{q}'(s)^T \mathbf{q}'(s) \dot s ^2 \leq v^2_{max}, \\
	-a_{max}\leq (\mathbf{q(s)}''\dot s^2 + \mathbf{q(s)}' \ddot s)_{\rm \{x,y,z\}} \leq a_{max},
	\end{gather}
\end{subequations}
which are then applied at each discrete position $s_i$.
Substituting (\ref{eq:uixi}) into (\ref{eq:qscons}) leads to linear constraints on $x_i$ and $u_i$, expressed as
\begin{equation}
\label{equ:constraints}
\mathscr{C}_i:= a_i u_i + b_i x_i + c_i \leq 0,
\end{equation}
where $a$, $b$ and $c$ are the coefficients of the linear expression. For (10a), $a=0, b=\mathbf{q}'(s)^T \mathbf{q}'(s), c = -v^2_{max}$; for (10b), $a=\pm \mathbf{q(s)}'_{\rm \{x,y,z\}}, b=\pm \mathbf{q(s)}''_{\rm \{x,y,z\}}, c= a_{max}$.
The $i$-th stage sets of admissible states and controls are defined as
\begin{equation}
	\mathcal{X}_i := \{x|\exists u: (u,x)\in \mathscr{C}_i\}, 
	~\mathcal{U}_i(x) := \{u|(u,x)\in \mathscr{C}_i\}.
\end{equation}

This process is shown in Fig. \ref{pic:path_para}, Step 2, but note that this sub-figure is just an inaccurate visualization of constraints formulation. Because we don't explicitly formulate any constraints of $\dot{s}^2$ (i.e., $x$) but only formulate constraints of $x$ and $u$ together in (\ref{equ:constraints}), which are then directly incorporated into linear programming in (\ref{eq:backward}b) of Step 3.

\subsubsection{\textbf{Controllable Sets Computation} ~(backward)} 
Given a set of states $\mathbb{I}$, where $x_{i+1}\in\mathbb{I}_{i+1}$, we can determine the lower and upper bounds $(x^-,x^+)$ for an admissible state $x_i$ by solving the following two Linear Programs:
\begin{subequations}
	\label{eq:backward}
	\begin{align}
	x^- := \min_{x, u}~x,~~~x^+ := \max_{x, u}~x,\\
	\textup{s.t.}~ (u_i, x_i)\in \mathscr C_i, ~ x+2 \Delta_i u \in \mathbb{I}_{i+1}.
	\end{align}
\end{subequations}
The interval $(x^-,x^+)$ is referred to as the \textit{one-step set} $\mathcal{Q}_i(\mathbb{I}_{i+1})$, signifying the existence of a state $\tilde{x}\in\mathbb{I}_{i+1}$ and an admissible control $u\in\mathcal{U}_i$ that guides the system from $x \in \mathcal{Q}_i$ to $\tilde{x}$.

As illustrated in Fig. \ref{pic:path_para}, Step 3, starting with $\dot s_N = \lVert\dot{\mathbf{q}}\lVert / \lVert\mathbf{q}'\lVert$ \textit{$i$-stage controllable set} $\mathcal{K}_i(\mathbb{I}_N)$ can be iteratively computed as follows:
\begin{equation}
\mathcal{K}_N(\mathbb{I}_N)=\dot{s}_N^2,~
\mathcal{K}_i(\mathbb{I}_N)=\mathcal{Q}_i(\mathcal{K}_{i+1}(\mathbb{I}_N)).
\end{equation}
The derived \textit{$i$-stage controllable set} $\mathcal{K}_i(\mathbb{I}_N)$ implies the existence of a state $x_N\in\mathbb{I}_N$ and a sequence of admissible controls $u_i,...,u_{N-1}$ that guide the system from $x \in \mathcal{K}_i$ to $x_N$.

Throughout this process, if $\mathcal{K}_i(\mathbb{I}_N) = \emptyset$ or $x_0 \notin \mathcal{K}_0$, it indicates that the path $\mathbf{q}(s)$ cannot be time-parameterized under such boundary conditions and dynamic constraints, and thus should be excluded from the primitive library.

\subsubsection{\textbf{Optimal Controls Construction }(forward)} 
In this procedure, the optimal states and controls are determined in a \textit{greedy} manner: at each stage $i$, the highest admissible control $u$ is chosen if it results in the next state falling within the $(i + 1)$-stage controllable set. The greedy strategy is reasonable because $u$ defined in (\ref{eq:uixi}) indicates the growth rate of the trajectory progress parameter $s$ which is monotonically increasing with respect to time $t$.

Given a state $x_i^*$, the optimal control $u_i^*$ is obtained by solving the following 
\begin{subequations}
	\label{eq:forward}
	\begin{gather}
	u_i^* = \argmax_{u}~u,\\
	\textup{s.t.}~u\in\mathcal{U}_i(x_i), 
	x_i^*+2 \Delta_i u \in \mathcal{K}_{i+1}.
	\end{gather}
\end{subequations}
With the known start velocity of the path $x_0^*=\dot{s}_0^2$, we can iteratively compute the optimal states for each stage using the above method, as depicted in Fig. \ref{pic:path_para}, Step 4.

The optimal time allocation ${t}_{0:N}$ is then determined by
\begin{subequations}
	\label{eq:t}
	\begin{gather}
		\dot{s}_{average}=(\sqrt{x^*_i} + \sqrt{x^*_{i+1}})/2,\\
		\Delta_t = (s_{i+1} - s_i)/\dot{s}_{average},\\
		t_{i+1} = t_i + \Delta_t.
	\end{gather}
\end{subequations}
%by the distances $\Delta_i$ and the average speeds of each segment $\dot{s}_{average}=(\sqrt{x^*_i} + \sqrt{x^*_{i+1}})/2$. 
Subsequently, a geometric path is parameterized by the time set $t_{0:N}$ to obtain a time-optimal and dynamically feasible trajectory $\mathbf{q}(s(t))$. Only trajectories that are parametrically feasible are included in the motion primitive library $\mathcal{P}$.

As constraints are imposed on a finite number of discrete positions, it is essential to analyze satisfaction errors. The high-accuracy interpolation scheme introduces a satisfaction error of $O(\Delta^2)$, which is influenced by the number of discrete positions $N$. Setting $N = 1000$ has proven to yield satisfactory solution accuracy. For a more detailed exploration of the satisfaction error, please refer to \cite{pham2018new}. The Seidel's LP algorithm \cite{seidel1991small} is employed to solve the linear program problems in (\ref{eq:backward}) and (\ref{eq:forward}), providing exact solutions. 
All computations are performed offline to compose a time-optimal and dynamically feasible motion primitive library.

\section{Fast Collision Checking}
\label{sec:collision_check}

\begin{figure}[t]
	\centering
	\includegraphics[width=0.9\linewidth]{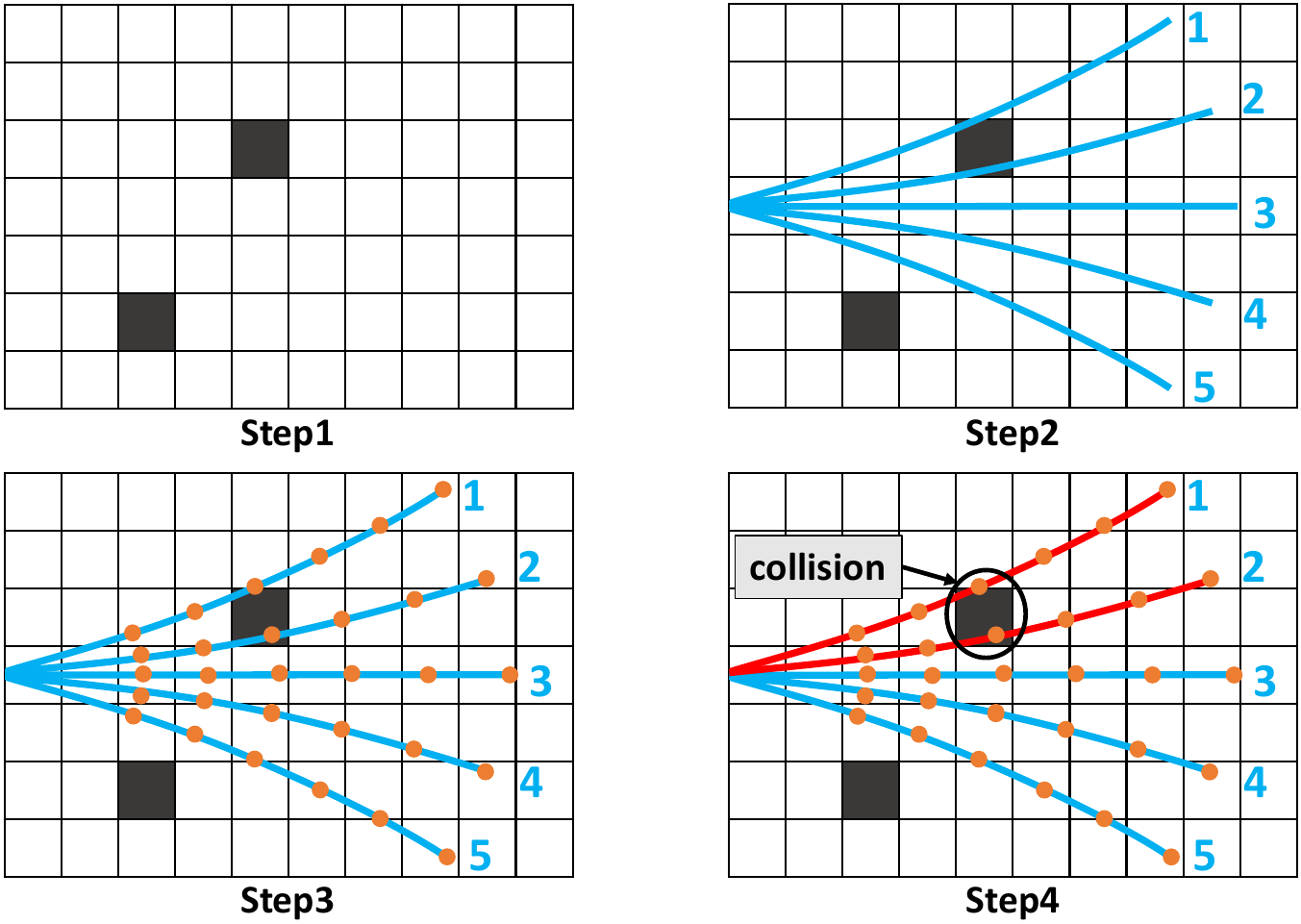}
        \caption{Illustration of typical robot-obstacle collision checking methods \cite{ryll2019efficient, yang2021intention, collins2020efficient, florence2020integrated} depicted in four distinct steps. Obstacles are depicted as dark boxes. Primitives are represented by blue curves, and their corresponding sample points are indicated as orange dots. Unsafe primitives are highlighted in red.}
	\label{pic:tradtion_check}
	 \vspace{-0.5cm}
\end{figure}

As the scale of swarm grows and the density of obstacles intensifies, the environment becomes increasingly fragmented. 
This fragmentation necessitates the planner to thoroughly explore the accessible free space, resulting in a rapid escalation of collision checks. 
Consequently, the adoption of an efficient collision checking method becomes crucial for ensuring swift and reliable responses to complex environments.
However, traditional methods \cite{ryll2019efficient, yang2021intention, collins2020efficient, florence2020integrated} adopt piece-by-piece, sample-by-sample approaches for robot-obstacle collision checking.
As shown in Fig. \ref{pic:tradtion_check}, these methods frequently repeat the query of the interaction between the robot and obstacles, making the time complexity dependent on the number of primitives multiplied by the number of sample points for each primitive.
When addressing robot-robot collision checking, a similar mechanism is applied, resulting in a time complexity estimated at $O(mnh)$, where $m,n,h$ represent the primitive number, sample number, and swarm scale, respectively.

To surmount these challenges and ensure effective spatial coverage and scalability within the swarm system, we draw inspiration from the work by \cite{zhang2020falco} and present a novel approach to expedited collision checking for swarm planning. Zhang et al. \cite{zhang2020falco} proposed an efficient way for collision checking by detecting and storing the occupancy information offline and then the collision checking module can perform efficient table queries only in flight. We extended this method by designing a mechanism for robot-robot conflict detecting which should consider not only static obstacles, but the movements of nearby robots as well.

\subsection{Offline Generated Spatial and Spatio-Temporal Occupancy Relationships}

\begin{figure}[t]
    \centering
    \includegraphics[width=0.9\linewidth]{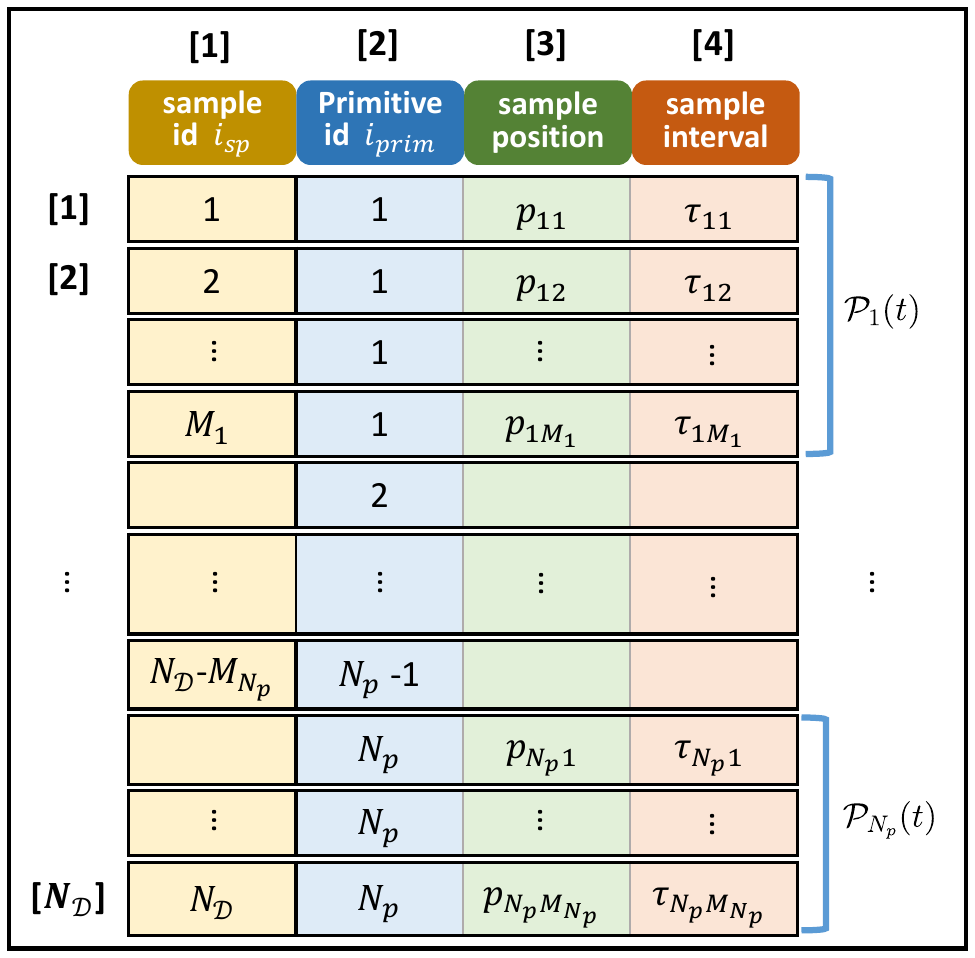}
     \caption{Visualization of the 2-D table $\mathcal{D}$ that encodes index from the primitives $\mathcal{P}_i(t)$, where $i\in\{1,2,...N_p\}$. The numbers in brackets represent row and column indices. The sample number for each primitive is denoted as $M_{i}$, $i\in\{1,2,...N_p\}$. Because the optimized time allocation for each primitive varies, the number of sample points $M_{i}$ also varies based on the same time resolution $t_{res}$. Sample interval $\tau_{ij} := [t_{start}, t_{end}]_{ij}$, $i\in\{1,2,...N_p\}$, $j\in\{1,2,...M_{i}\}$, $t_{end}-t_{start}=t_{res}$. Sample positions are sampled at the middle point of the time interval: $\mathbf{\textit{p}}_{ij} = \mathcal{P}_i((t_{start,ij}+t_{end,ij})/2)$. The total number of entries in the table is size($\mathcal{D}$) = $N_\mathcal{D}=\sum_{i=1}^{N_p} M_{i}$.}
    \label{pic:data_structure}
    \vspace{-0.5cm}
\end{figure}

\begin{algorithm}
    \begin{algorithmic}[1]
        \REQUIRE $\mathcal{P}(t)$, $s_{res}$, $t_{res}$, $d_1$, $d_2$
        \ENSURE $\mathcal{R}_o$, $\mathcal{R}_t$
        
        /* $\mathcal{G}, \mathcal{D}, \mathcal{T}$ \textbf{for Occupancy Relationships} */
        \STATE $\mathcal{G}$ = \textit{SplitCoverageSpace}($\mathcal{P}(t)$, $s_{res}$)
	\STATE $\mathcal{D}$ = \textit{DiscretePrimitives}($\mathcal{P}(t)$, $t_{res}$)
	\STATE $\mathcal{T}$ = \textit{BuildKDTree}($\mathcal{D}.$column(3))
        
	/* \textbf{$\mathcal{R}_o$ for robot-obstacle collision checking} */
        \STATE $\mathcal{I}_{D1}$ = $\mathcal{T}$.\textit{QueryBallPoint}($\mathcal{G}$, $d_1$)
        \STATE $\mathcal{R}_o$.allocate(size$(\mathcal{G})$)
        \FOR{$i$ from $1$ to size$(\mathcal{G}$)}
            \STATE $\mathcal{I}_{tmp}$ = SortByID($\mathcal{I}_{D1}[i]$), $last\_id$ = $0$
            \FOR{$i_{sp}$ from 1 to size$(\mathcal{I}_{tmp})$}
                \STATE $i_{prim}$ = $\mathcal{D}[i_{sp}, 2]$  % Assuming you're accessing column 1
                \IF{$i_{prim}$ == $last\_id$}
                    \STATE \textbf{continue}
                \ENDIF
                \STATE $\mathcal{R}_o[i]$.push\_back($i_{prim}$)
                \STATE $last\_id$ = $i_{prim}$
            \ENDFOR
        \ENDFOR
	
	/* \textbf{$\mathcal{R}_t$ for robot-robot collision checking} */
	\STATE $\mathcal{I}_{D2}$ = $\mathcal{T}$.\textit{QueryBallPoint}($\mathcal{G}$, $d_2$)
	\STATE $\mathcal{R}_t$.allocate(size$(\mathcal{G})$)
        \FOR{$i$ from $1$ to size$(\mathcal{G})$}
            \STATE $\mathcal{I}_{tmp}$ = SortByID($\mathcal{I}_{D2}[i]$), $last\_id = 0$
            \FOR{$i_{sp}$ from 1 to size$(\mathcal{I}_{tmp})$}
                \STATE $i_{prim}$ = $\mathcal{D}[i_{sp}, 2]$, $\tau=\mathcal{D}[i_{sp}, 4]$
                \IF{$i_{prim}$ != $last\_id$}
                    \STATE $\mathcal{R}_t[i]$.push\_back($i_{prim}$, $\tau.t_{start}$, $\tau.t_{end}$)
                \ELSE
                    \STATE $\mathcal{R}_t[i]$.back().$t_{end}$ = $\tau.t_{end}$
                \ENDIF
                \STATE $last\_id$ = $i_{prim}$
            \ENDFOR
        \ENDFOR
        \RETURN $\mathcal{R}_o$, $\mathcal{R}_t$
    \end{algorithmic}
\caption{Offline Occupancy Relationships}
\label{alg:st_mpl}
\end{algorithm}

To improve the collision checking efficiency for motion primitives, our method establishes \textit{offline spatial and spatio-temporal occupancy relationships} by pre-computing an index system that stores the occupancy status between primitives $\mathcal{P}(t)$ and the surrounding space.
The foundation of this index system involves a twofold process: discretizing the spatial space into grids and subsequently associating motion primitives passing through each grid.

Firstly, the continuous space is discretized into distinct grids.
For each grid, a search operation is conducted to identify and categorize motion primitives that traverse the grid's vicinity within a predefined radius $d$.
This implies that there will be a collision between the motion primitive and the grid at some point in time.
For those primitives that have been checked for conflicts, both the primitive ID and the time interval during which the primitive enters and exits the search radius to the grid are recorded.
We refer to this record as \textit{spatial and spatio-temporal occupancy relationships}, which is employed to index infeasible primitives from unsafe environmental grids.

The procedure for constructing occupancy relationships is detailed in Algorithm \ref{alg:st_mpl}.
% with some key data structures visualized Fig. \ref{pic:data_structure}. 
The inputs include the motion primitive library $\mathcal{P}(t)$, the spatial resolution $s_{res} \in \mathbb{R}^+$ to discretize the space, the time step $t_{res} \in \mathbb{R}^+$ of sampling $\mathcal{P}(t)$, the radius $d \in \mathbb{R}^+$ including $d_1$ and $d_2$ indicating the required clearance to avoid obstacles and other robots, respectively.
The outputs include spatial occupancy relationship $\mathcal{R}_o$, denoting grid-primitive associations, and spatio-temporal occupancy relationship $\mathcal{R}_t$, signifying grid-primitive-time interval associations.

In the initial step (Line 1), the function \textit{SplitCoverageSpace} computes the set of all grids $\mathcal{G}:= \{\mathbf{g}_1, \mathbf{g}_2, \cdots\}, \mathbf{g}_i \in \mathbb{R}^3$ that are inside the axis-aligned bounding box of spatial resolution of $s_{res}$ traversed by at least one motion primitive in $\mathcal{P}(t)$.
The computational complexity is $O(m)$ with $m$ represents the primitive number because the boundary of $\mathcal{G}$ is only obtained at the terminal points of each motion primitive.
Function \textit{DiscretePrimitives} (Line 2) outputs $\mathcal{D}$, a 2-D table visualized in Fig. \ref{pic:data_structure} that encodes sampled points from $\mathcal{P}(t)$ with time step $t_{res}$ and can be indexed using the notation $\mathcal{D}[row~id, column~id]$ with $id$ starts from 1. 
The computational complexity of constructing $\mathcal{D}$ is $O(\Sigma_{i=1}^{m}n_i)$ with $n_i$ the sample number of primitive $i$.
For precise collision detection, dense point sampling on each primitive is essential. 
While increasing the number of primitives and point samples enhances accuracy, it also escalates time complexity.
To strike a balance, we use the sample position in $\mathcal{D}$ to build a class of kd-tree $\mathcal{T}$ (Line 3) of complexity $O(kn\log n)$ \cite{brown2014building} with $n$ the total sample number and $k$ the degree of the kd-tree. 
$\mathcal{T}$ has a method $QueryBallPoint$ that finds sampled 3-D points $\mathbf{\textit{p}} \in \mathcal{D}$ that are within the radius $d$ from each grid center.
Function $QueryBallPoint$ returns the results to $\mathcal{I}_D := \{\mathbf{\textit{i}_{D0}}, \mathbf{\textit{i}_{D1}}, \cdots\}$ that satisfies size$(\mathcal{I}_D) =$ size$(\mathcal{G})$. 
The complexity of this function is $O({\rm size}(\mathcal{G}) \log n)$.

The function SortByID$(\mathcal{I}_D[i])$ (Line 7, 20; $O({\rm size}(\mathcal{I}_D[i])^2)$ complexity) sorts the rows of $\mathcal{I}_D[i]$ in increasing order of $i_{sp}$, ensuring that rows sampled from the same motion primitive are stacked together like $\mathcal{D}$.
The following code in the for-loop (Line 8-15, Line 21-29) stores the conflicted primitive ID $i_{prim}$ and the ID with the time interval $\{i_{prim}, t_{start}, t_{end}\}$ of each grid $\mathcal{G}$ into the algorithm output  $\mathcal{R}_o$, $\mathcal{R}_t$.
These represent the spatial occupancy relationship and spatial-temporal occupancy relationship, respectively.
Elements of $\mathcal{R}_o$ and $\mathcal{R}_t$ are retrieved with the notation $\mathcal{R}_o[i]$ and $\mathcal{R}_t[i]$, where $i \in \mathbb{N}$ denotes the index of each grid $\mathbf{g}_i$.

\subsection{Online Robot-obstacle Collision Checking}

\begin{figure}[t]
	\centering
	\includegraphics[width=1.0\linewidth]{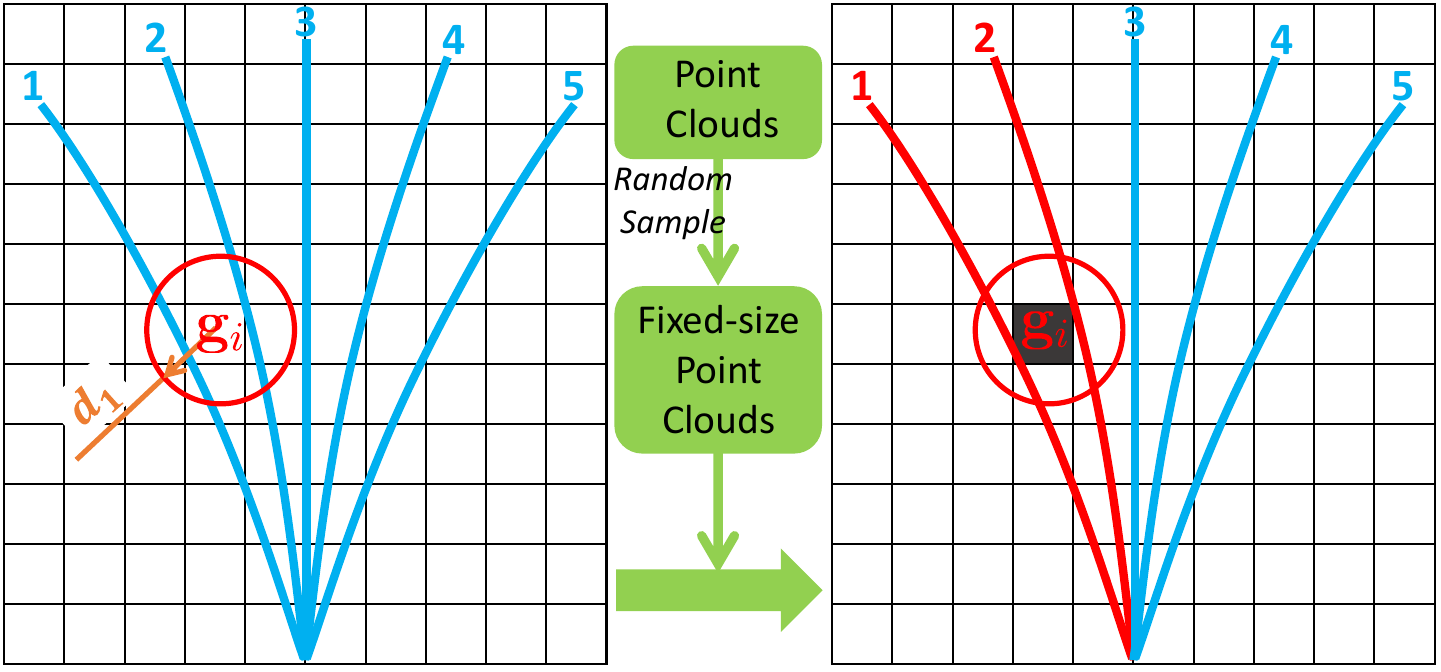}
        \caption{Robot-obstacle collision checking. The grid $\mathbf{g}_i$ associates primitives $1$ and $2$, i.e., $\mathcal{R}_o[i]=\{1,2\}$ according to Algorithm \ref{alg:st_mpl}. $d_1$ represents the query distance. The dark grid represents obstacles. The blue primitives are considered safe, while the red primitives are marked as unsafe.}
	\label{pic:obstacle_collision_check}
	\vspace{-0.4cm}
\end{figure}

In Fig. \ref{pic:obstacle_collision_check}, the spatial occupancy relationship $\mathcal{R}_o$ stores the grids $\mathcal{G}$ with its conflicted primitives. We set the query distance as 
\begin{equation}
d = d_1 = \frac{\sqrt{3}}{2} s_{res} + r_{infl},
\end{equation}
where $r_{infl}$ is a safe distance required between the drone and obstacles. 

Then for any obstacle part that falls within a grid of index $i$, we can retrieve the collided primitives from $\mathcal{R}_o[i]$ with $O(1)$ complexity. 
During real-world implementations, when the sensor receives the point clouds of obstacles, we employ the random sample method \cite{vitter1984faster} to obtain fixed-number $N_{pc}$ of point clouds while ensuring the fidelity of environments.
We then project these point clouds onto the grids $\mathcal{G}$, and the associated primitives are batch-labeled as unsafe. 
The time consumption of the robot-obstacle collision checking is only related to the number of point clouds and remains independent of the number of primitives.
The proposed method reduces the number of point clouds to a fixed number $N_{pc}$ so that collision checking can be finished quickly in deterministic time.
The time complexity of our method to process a frame of point cloud is $O(N_{pc})$. Furthermore, to ensure robot safety, typical methods \cite{zhou2020ego,zhou2021ego,zhou2022swarm} usually inflate a grid map online by a specific safety distance. In contrast, our method requires only an offline setting of the safe distance $r_{infl}$ without incurring any online computational cost on map inflation.

\subsection{Online Robot-robot Collision Checking}
\begin{figure}[t]
	\centering
	\includegraphics[width=1.0\linewidth]{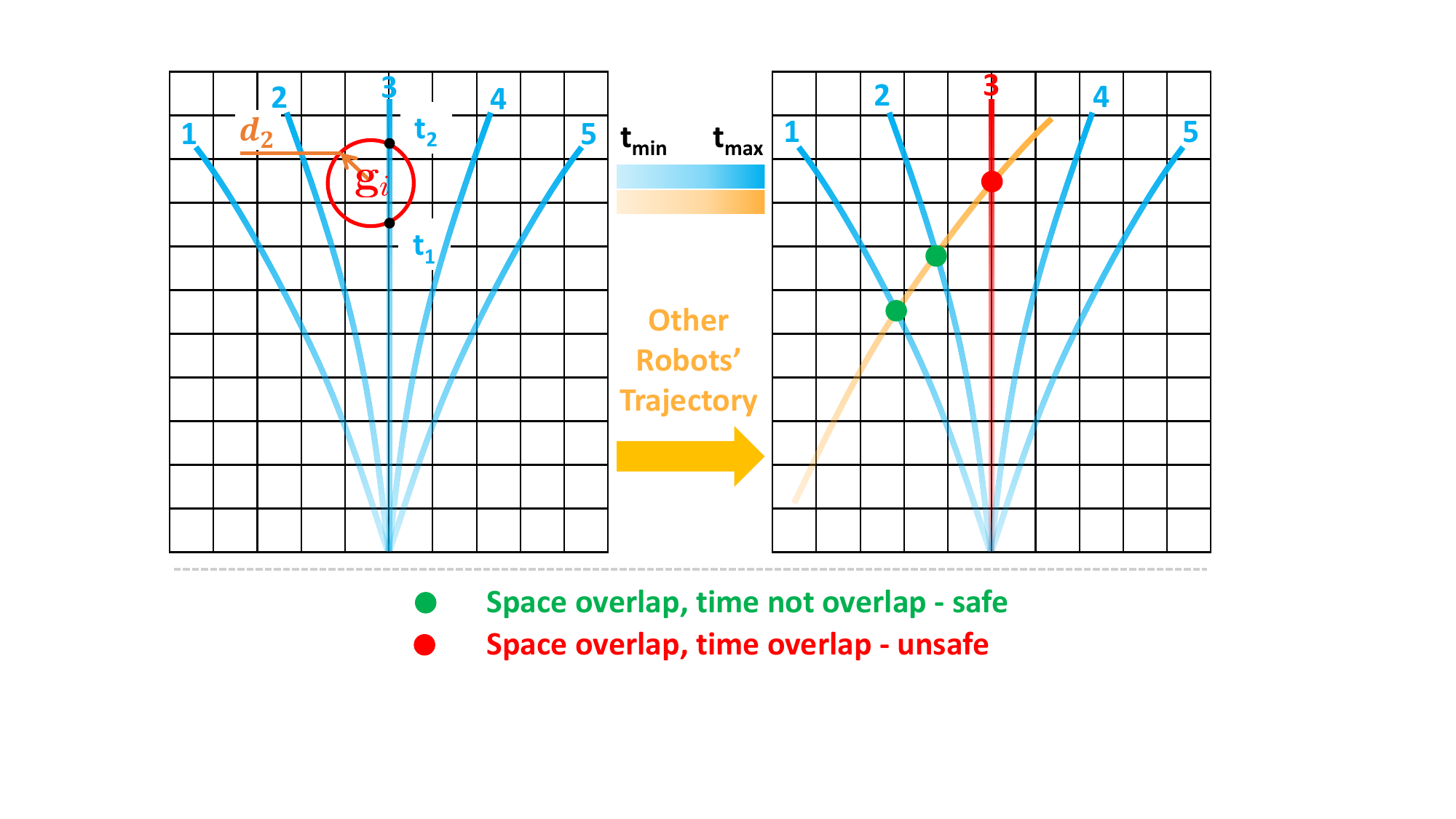}
        \caption{Robot-robot collision checking. The grid $\mathbf{g}_i$ corresponds to the time interval $[t_{1}, t_{2}]$ of primitive $3$, i.e., $\mathcal{R}_t[i]=\{\{3,t_{1}, t_{2}\}\}$ according to Algorithm \ref{alg:st_mpl}. $d_2$ denotes the query distance. The yellow primitive represents the trajectory of another robot. The blue primitives are considered safe, while red primitive is marked as unsafe.}
	\label{pic:agent_collision_check}
	\vspace{-0.4cm}
\end{figure}

\begin{figure*}[t]
	\centering
	\includegraphics[width=1.0\linewidth]{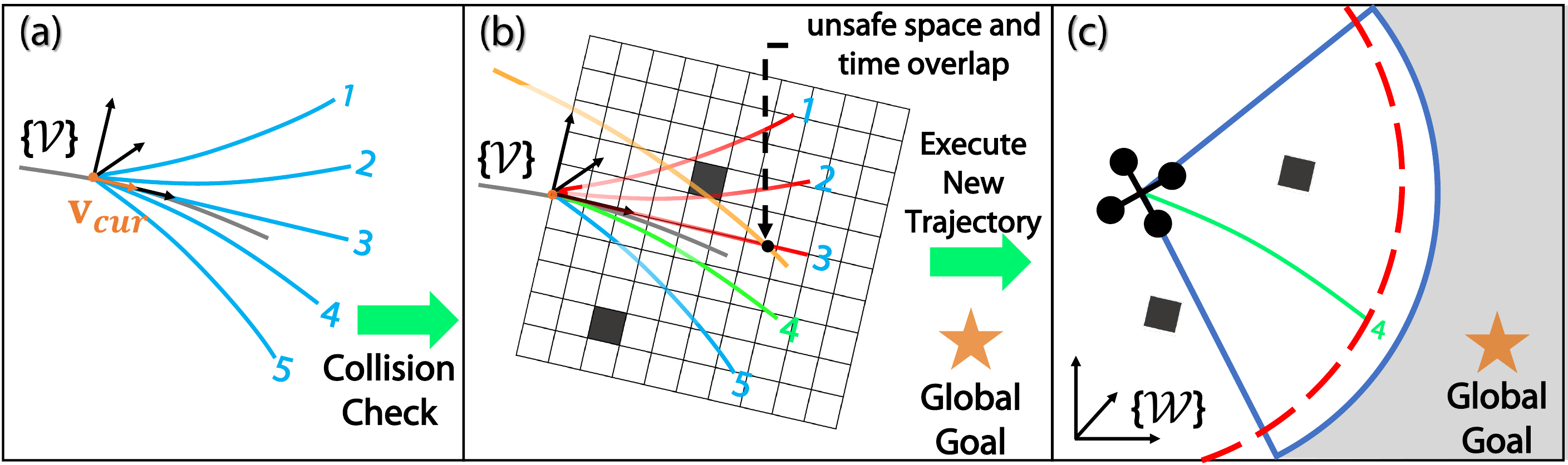}
        \caption{Online local replanning. (a) Velocity-aligned coordinate system $\mathcal{V}$. (b) Trajectory selection. Dark grids represent obstacles, and the yellow curve shows another robot's trajectory. Decreasing transparency indicates progressing time on the trajectory. The orange star marks the global goal. Primitive $4$ is chosen as the new trajectory with the lowest cost which is defined in Section~\ref{sec:traj_select}. (c) Receding horizon strategy. The red dashed line depicts the planning horizon, while the blue sector represents the robot's perception range. The light gray region signifies unexplored territory.}
	\label{pic:replan}
	\vspace{-0.6cm}
\end{figure*}

In Fig \ref{pic:agent_collision_check}, the spatio-temporal occupancy relationship $\mathcal{R}_t$ stores the ID and time interval of the conflicted primitive with each grid in $\mathcal{G}$. 
We set the query distance as follows
\begin{equation}
d = d_2 = \frac{\sqrt{3}}{2} s_{res} + 2 r_{robot},
\end{equation}
where $r_{robot}$ is the radius of robots. 
 
We communicate and receive the trajectories of other robots within two planning horizons (as defined in Fig. \ref{pic:replan}c), because robots beyond this distance are sure to be collision-free in the next trajectory planning.
Then we project these trajectories onto the grids $\mathcal{G}$ along with the time interval $[t_{start}, t_{end}]_l$ indicating when the drone $l$ will occupy that grid $\mathbf{g}_i$. 
For each motion primitive with ID $i_{prim}$ in $\mathcal{R}_t[i]$, the corresponding time interval $[t_{start}, t_{end}]_{i_{prim}, i}$ is retrieved and compared with $[t_{start}, t_{end}]_l$.
If $[t_{start}, t_{end}]_{i_{prim}, i} \cap [t_{start}, t_{end}]_l \neq \emptyset$, then $i_{prim}$-th primitive will collided with drone $l$ in the future, thus $i_{prim}$-th primitive is marked unsafe.
The time consumption of the robot-robot collision checking is only related to the number of trajectories of other robots. The algorithm complexity of our method is $O(N_d)$ with $N_d$ the number of drones within the two planning horizons. 

In summary, the proposed method minimizes unnecessary online computation costs for collision checking and operates directly on raw point clouds without any computationally expensive environment representation. Its time complexity is independent of the total number of primitives, as quantitatively demonstrated in Section \ref{sec:eva_check}. Thus in an onboard processor, the number of primitives can be set sufficiently high to provide high-quality candidates.

\section{Online Local Replanning}
\label{sec:re_plan}

To react to incremental environmental sensing in unknown surroundings, the planner repeatedly generates trajectories. Trajectory replanning is triggered based on two criteria: (1)~reaching a specified time threshold, or (2)~detecting any collision along the current executing trajectory.

\subsection{Velocity-Aligned Coordinate System}
\label{sec:vcs}

As introduced in Sec. \ref{subsubsec:lib}, for any trajectory in the motion primitive library $\mathcal{P}$, the start position ${\mathbf{q}}(s_0) \equiv \mathbf{0}$ and direction of the velocity $\dot{\mathbf{q}}(s_0)$ is tangent to the x-axis. 
Therefore, to ensure trajectory continuity, we have to align the direction of the start velocity $\dot{\mathbf{q}}(s_0)$ with the direction of robot's present velocity $\mathbf{v}_{cur}$. To achieve this, a velocity-aligned coordinate system $\mathcal{V}$ is established based on $\mathbf{v}_{cur}$, depicted in Fig. \ref{pic:replan}a. This system is defined as follows:
\begin{equation}
\begin{gathered}
x_{axis} = \frac{\mathbf{v}_{cur}}{\lVert \mathbf{v}_{cur} \rVert} ,\\
y_{axis} = x_{axis} \times \mathbf{g},\\
z_{axis} = x_{axis} \times y_{axis},
\end{gathered}
\end{equation}
where $x_{axis}, y_{axis}, z_{axis}$ represent the axes of $\mathcal{V}$, while $\mathbf{g}$ symbolizes the normalized gravity vector. All primitives share a common origin in $\mathcal{V}$ and are tangential to the x-axis at their respective starting velocity. The selection of start velocities for all primitives is determined by the magnitude of the current velocity $\lVert\mathbf{v}_{cur}\rVert$.

The transformation $T_\mathcal{V}^\mathcal{W}$ from the velocity-aligned coordinate system $\mathcal{V}$ to the world frame $\mathcal{W}$ is defined as follows:
\begin{equation}
R_\mathcal{V}^\mathcal{W} = [x_{axis},~y_{axis},~z_{axis}],\\
\end{equation}
\begin{equation}
T_\mathcal{V}^\mathcal{W} = \left[
\begin{matrix}R_\mathcal{V}^\mathcal{W} & \mathbf{p}_{start}\\
\textbf{0}^T & 1
\end{matrix}
\right]
\end{equation}
where $R_\mathcal{V}^\mathcal{W}$ is a rotation matrix, and $\mathbf{p}_{start}$ represents the robot's position within the world coordinate system $\mathcal{W}$.

In practice, we perform a transformation of environmental information, such as obstacles and other trajectories, from the world coordinate system $\mathcal{W}$ to the velocity frame $\mathcal{V}$, which corresponds to the coordinate frame of the primitive library.

\subsection{Trajectory Selection}
\label{sec:traj_select}

In Fig. \ref{pic:replan}b, we begin by excluding unsafe primitives from collision checking. Subsequently, we formulate a cost function $\mathcal{C}$ for each safe primitive based on user-defined specifications. The cost function $\mathcal{C}$ is expressed as follows:
\begin{equation}
	\label{eq:cost}
	\begin{gathered}
	\mathcal{C} = \lambda_g \mathcal{C}_{goal} + \lambda_b \mathcal{C}_{bound}, ~ where\\
	\mathcal{C}_{goal} = \lVert \mathbf{p}_{end} - \mathbf{p}_{goal} \rVert - \lVert \mathbf{p}_{start} - \mathbf{p}_{goal} \rVert,\\
	\mathcal{C}_{bound} = \begin{cases}
	C_\mathcal{B}, & \text{if}~\mathbf{p}_{end} \notin \mathcal{B}\\
	0, & \text{if}~\mathbf{p}_{end} \in \mathcal{B}
	\end{cases},
	\end{gathered}
\end{equation}
where, $\lambda_g$ and $\lambda_b$ represent corresponding weights, and $\mathbf{p}_{start}$ denotes the robot's position in the world coordinate system $\mathcal{W}$, $C_\mathcal{B} > 0$ is a constant scalar. The goal cost $\mathcal{C}_{goal}$ aims to maximize the improvement of getting close to the $\mathbf{p}_{goal}$. The boundary cost $\mathcal{C}_{bound}$ penalizes end positions $\mathbf{p}_{end}$ outside the allowed bounds $\mathcal{B}$, like some user-defined flyable areas.

The proposed approach can also accommodate other desired criteria by extending the cost function $\mathcal{C}$, such as yaw angle change rate. Finally, we select the primitive with the minimum cost as the new trajectory to be executed.

\subsection{Receding Horizon Strategy}

We employ a receding horizon strategy for the implementation of replanning in autonomous swarm navigation, illustrated in Fig. \ref{pic:replan}c. When the conditions for replanning are met, the planner selects an optimal local trajectory to execute as in Section \ref{sec:traj_select}. 
The replanning process is reiterated until the robots successfully reach their global goals.
The local trajectory is planned within a planning horizon, which is determined by the robot's perception range. This planning horizon confines the scope of trajectory calculation. Moreover, the initiation of replanning for robots is carried out asynchronously. This strategic approach empowers robots to promptly respond to dynamic environment changes and conflicts with other robots. Simultaneously, the long-term planning to achieve their global goals is upheld.

\section{Implementation Details}
\label{sec:details}

%\subsection{System Setup}
%\label{sec:System_Setup}

For simulations, our simulator encompasses a quadrotor dynamics model, a random map generator, a simulated depth sensor, and communication module. 
All the simulations share the same parameter configuration for trajectory planning, except for the number of robots, their corresponding start positions and goals, and obstacles placed according to different experiment requirements.
With the exception of the large-scale simulations (Section \ref{sec:large_scale}), all simulations and comparisons were conducted on a personal computer with an Intel Core i7 10700 CPU (8 cores, 16 threads), an NVIDIA GeForce GTX 1660 Ti GPU, and 32 GB RAM.
For the large-scale simulation of swarms, we utilized the ecs.hfg6.20xlarge workstation\footnote{\url{https://www.alibabacloud.com/solutions/sap?spm=a3c0i.23458820.2359477120.13.61d67d3fDISyrx}}, which is further detailed in Section \ref{sec:large_scale}.
Notably, all drones were operated within independent parallel threads to maintain a decentralized and asynchronous system architecture.
In our real-world experiments, we utilize an open-source quadrotor platform\cite{zhou2022swarm}.

\begin{figure}[t]
	\centering
	\includegraphics[width=1.0\linewidth]{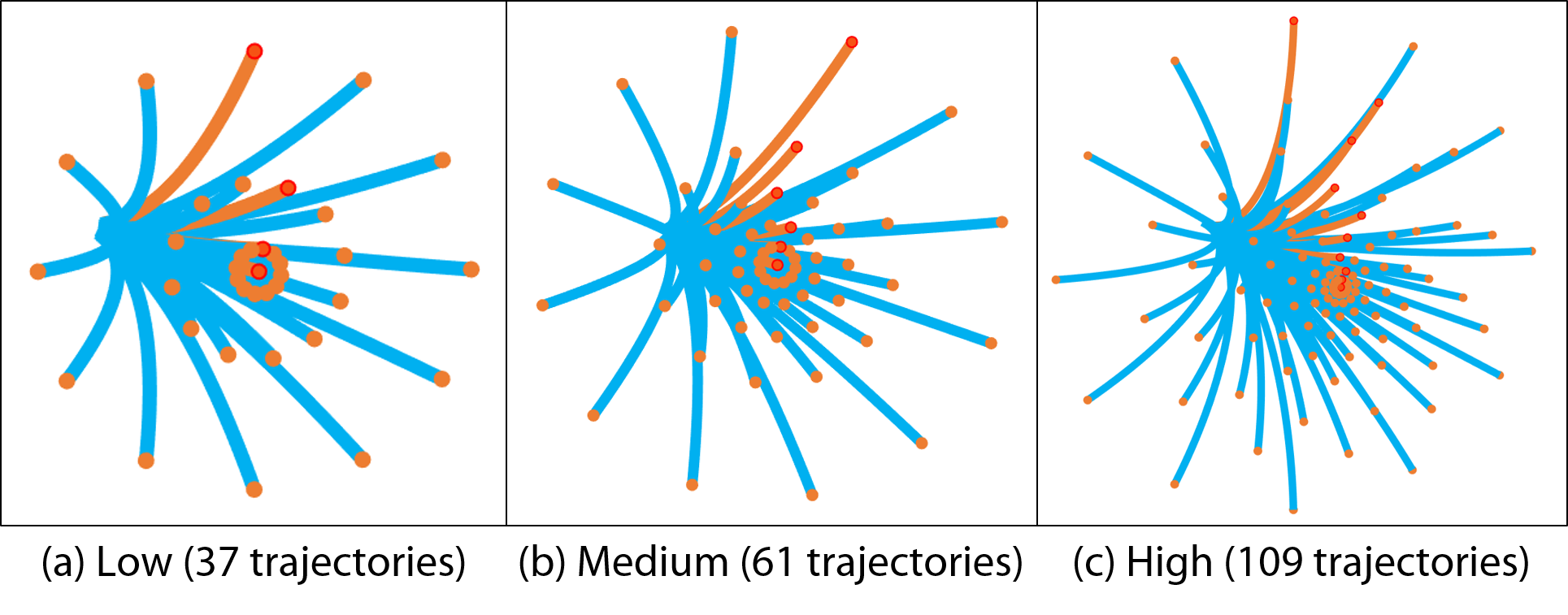}
	\caption{Three motion primitive libraries. (a) $r\in\{8, 20, 78, \infty\}m$; (b) $r\in\{6, 12, 20, 36, 78, \infty\}m$; (c) $r \in \{2, 3, 4, 6, 8, 12, 20, 36, 78, \infty\}m$; In the all three figures, $l=3m$, $\theta = \{0^\circ, -10^\circ, -20^\circ, 0^\circ ... , 0^\circ\}$, $D_{angle}=30^\circ$.} 
	\label{pic:three_motion_lib}
	\vspace{-0.2cm}
\end{figure}

\begin{figure}[t]
	\centering
	\includegraphics[width=1.0\linewidth]{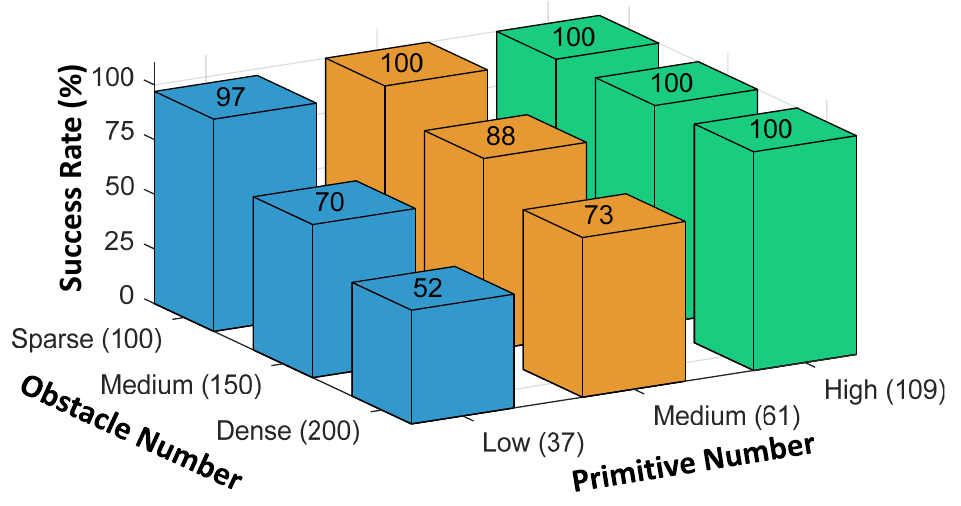}
	\caption{Success rate (\%) in different configurations.}
	\label{pic:success_rate}
	 \vspace{-0.5cm}
\end{figure}

\begin{figure*}[t] 
	\centering
	\includegraphics[width=1.0\linewidth]{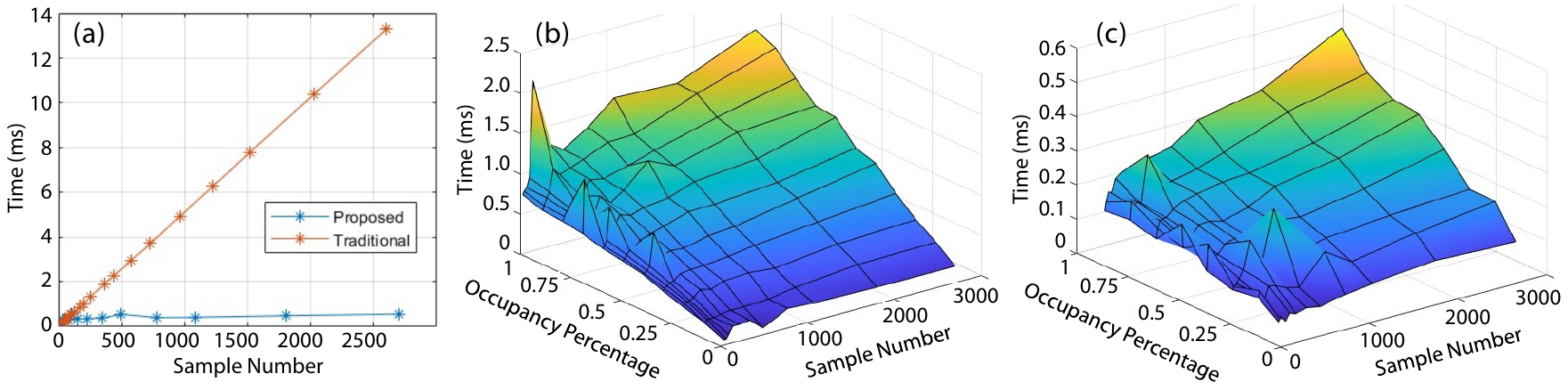}
	\caption{The computation time for checking robot-obstacle collisions. }
	\label{pic:obs_collision_time}
	 \vspace{-0.5cm}
\end{figure*}

\section{Evaluation}
\label{sec:evaluation}

In this section, we validate the proposed method through simulations and real-world experiments. 

\subsection{Success Rate Analysis of Motion Primitive Library}
\label{sec:per_analysis}

The success rate of autonomous navigation is evaluated in different scenarios. We utilize three motion primitive libraries with varying numbers of trajectories $N_{t}\in\{37, 61, 109\}$ (as shown in Fig. \ref{pic:three_motion_lib}) and three environments with different numbers of obstacles $N_{obs}\in\{100, 150, 200\}$. The cylindrical obstacles have an average radius of $0.6~m$, distributed in a $26\times20\times3~m$ space (see Fig. \ref{pic:comparison_single}).
In each of the nine scenarios, one drone starts from a random point  ($x=-18~m$, $y\in[-9,+9]~m$, $z=1~m$) at one side of the map and navigates to a goal position ($x=18~m$, $y\in[-9,+9]~m$, $z=1~m$) and repeats for 100 times. The success rate is recorded and presented in Fig. \ref{pic:success_rate}. 

From the results, the motion primitive library with a low number of trajectories performs poorly in dense environments due to its limited flexibility and difficulty in fitting complex trajectories. However, by increasing the primitive number in the motion primitive library to enhance flexibility, our planner demonstrates robust performance in all environments. Therefore, the proposed method is capable of handling autonomous navigation tasks by designing a reasonable primitive number for primitive library in complex and dense environments. In later tests, we mostly increase the primitive number to $181$ to further increase the possibility of finding better trajectories.

\begin{figure}[!h]
	\centering
	\includegraphics[width=0.9\linewidth]{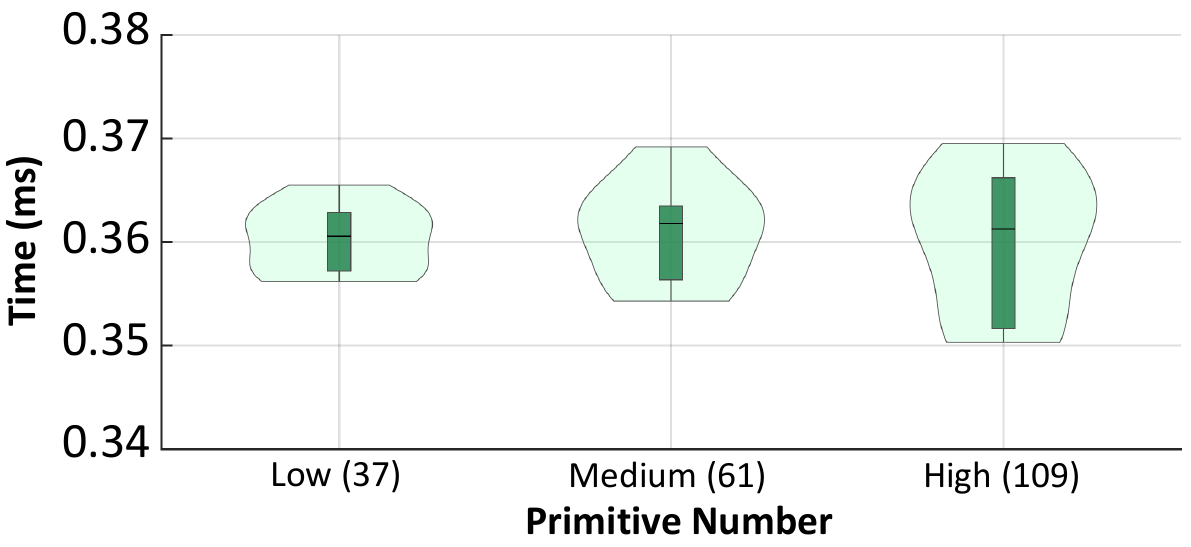}
	\caption{The computation time for checking robot-robot collisions.}
	\label{pic:agent_collision_time}
	\vspace{-0.7cm}
\end{figure}

\subsection{Fast Collision Checking}
\label{sec:eva_check}

We evaluate the efficiency of the proposed collision checking method in two experiments.

\begin{figure*}[!h]
	\centering
	\includegraphics[width=1.0\linewidth]{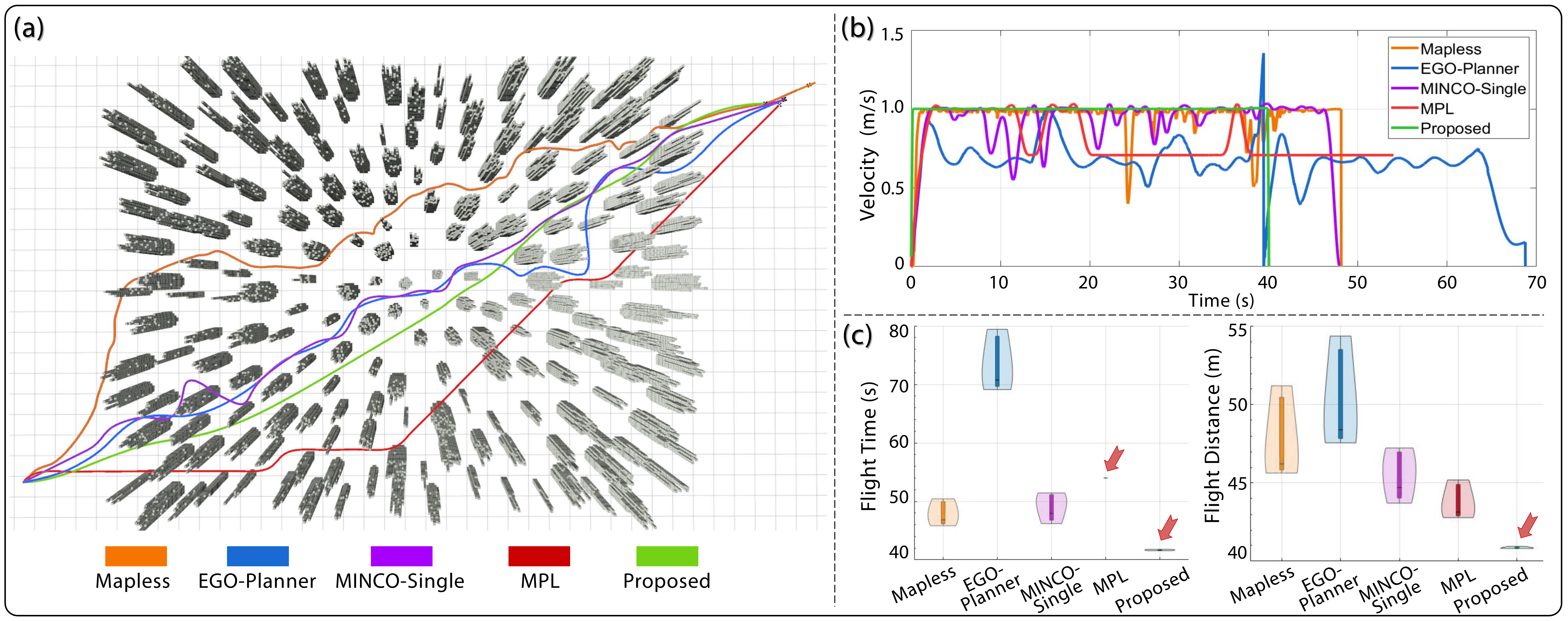}
	\caption{Single-robot comparison. (a) Visualization of executed trajectories. The drone starts from the same initial and goal positions in all methods. The gray pillars represent obstacles, and the colored curves depict the executed trajectories of each method. (b) Velocity profile comparison. (c) Metrics comparisons using violin graphs. \textit{Flight Time} and \textit{Flight Distance} are metrics used to evaluate the drone's performance in reaching the global goal.}
	\label{pic:comparison_single}
		\vspace{-0.1cm}
\end{figure*}

1) Robot-obstacle collision checking: In Sec. \ref{sec:collision_check} we discussed that the proposed method scales better in primitive number than the typical collision checking method illustrated in Fig. \ref{pic:tradtion_check}. This point is validated in Fig. \ref{pic:obs_collision_time}a. For the typical method, we generate discrete acceleration commands and impose them onto a system with zero initial position and velocity, leading to different trajectories. Then each trajectory is truncated to the same length of the proposed primitive library. Collision checking is then performed with the same map and the time to compute is recorded in Fig. \ref{pic:obs_collision_time}a. As we analyzed, the computation time of the traditional method grows linearly with the number of trajectories, while the proposed method shows a slow increase, which means a higher scalability.

In fact, the time complexity of the proposed method is linear to the number of point cloud, as we check collision for each obstacle point. Therefore we conduct another two tests in Fig. \ref{pic:obs_collision_time}b and c. 
The \textit{Occupancy Ratio} indicates the percentage of obstacles in space under the given map resolution. Generally speaking, areas with an occupancy ratio exceeding 30\% will be difficult to navigate. For reference, this ratio in Fig. \ref{pic:comparison_single} is approximately 15\%. From Fig. \ref{pic:obs_collision_time}b and c, the computation time increases linearly with the occupancy ratio, and the more primitives there are, the faster the growth. In addition, the resolution has a very significant impact on the computation time, as the point cloud density is directly proportional to the inverse of the resolution cubed. For the vast majority of scenes with an occupancy ratio less than 30\%, the collision checking time at a resolution of $0.1~m$ is usually less than $0.5~ms$ , and if the resolution is increased to $0.2~m$, the computation time will further decrease to below $0.2~ms$.

2) Robot-robot collision checking: We conducted experiments with $8$ drones evenly distributed in a circular empty space with a radius of $12~m$, as illustrated in Fig.~\ref{pic:bench_circle}.
The drones exchanged positions using three different motion primitive libraries. Each scenario was run $20$ times, and the results are shown in Fig. \ref{pic:agent_collision_time}. The collision checking time, averaged across all $8$ drones, consistently remained between $0.35$-$0.37ms$ in all three situations.

In summary, the proposed method enables rapid conflict checking without the need for computationally expensive environment representation or obstacle inflation. Its time complexity remains independent of the number of primitives, allowing a sufficiently high number of primitives to be used on an onboard processor, providing high-quality candidates.

\subsection{Single-robot Simulation and Comparison}
\label{sec:single_comparison}

The proposed planner serves as a high-performance single-robot local planner when the number of robots is set to one. We compared the proposed method with four state-of-the-art local planners: Mapless (2021) \cite{ji2021mapless}, EGO-Planner (2020) \cite{zhou2020ego}, and MINCO-Single (2022, the single-robot version of MINCO-Swarm\cite{zhou2022swarm} which uses the trajectory representation called MINCO \cite{wang2022geometrically}). Mapless is a lightweight approach using a kd-tree data structure. EGO-Planner and MINCO-Single are online optimization methods with an inflated map. Except that, MPL (2018, Motion Primitive Library) \cite{liu2018towards} is compared as a representative sampling-based planner. All methods are open-sourced, and we use default parameters. 
The flight area contains $200$ obstacles distributed in a $26\times20\times3~m$ space. To ensure fairness, we test all methods with the same perception range (except for MPL which uses global map), initial position $[-18.0, -9.0, 1.0]~m$, and goal position $[18.0, 9.0, 1.0]~m$. The number of primitives $N_{t}$ is set to $181$ for the proposed method. 
The results are presented in Fig. \ref{pic:comparison_single}c. 
The executed trajectories of the five methods in a random map are illustrated in Fig. \ref{pic:comparison_single}a, and their velocity profiles are shown in Fig. \ref{pic:comparison_single}b. 
MPL as a base planning library only supports global planning, which means it plans the whole trajectory at a once. Therefore its computation time is significantly larger than other methods.

\begin{table*}[t]
	\centering
        \caption{Benchmark Comparisons in a $12~m$ radius circular empty space containing $8$ drones. Bold text with an underscore represents a better value. ``\textit{Emergency}" refers to whether there is a dangerous situation where the drone spacing is below 0. ``\textit{Real-Time}" refers to whether the trajectory can be timely planned during the flight based on the latest environment information. ``\textit{Decentralized}" means that each drone plans its own trajectory.}
	\label{tab:comparison_empty}
	\renewcommand{\arraystretch}{1.20}
	\setlength\tabcolsep{3pt}
	\resizebox{\linewidth}{!}{
		\begin{tabular}{|c|c|c|c|c|c|c|}
			\hline
			\textbf{Method} &
			\textbf{Flight Time (s)} &
			\textbf{Flight Distance (m)} &
			\textbf{Computation Time (ms)} &
			\textbf{Emergency?} &
			\textbf{Real-Time?} &
			\textbf{Decentralized?} \\ \hline
			\textbf{RBP(batch size=1)} &
			114 &
			30.006 &
			1735.610* &
			\multirow{2}{*}{\textbf{\underline{No}}} &
			\multirow{2}{*}{No} &
			\multirow{2}{*}{No} \\ \cline{1-4}
			\textbf{RBP(batch size=4)} &
			114 &
			30.006 &
			1756.800* &
			&
			&
			\\ \hline
			\textbf{MADER} &
			28.482 &
			24.128 &
			7.450 &
			Yes &
			\textbf{\underline{Yes}} &
			\textbf{\underline{Yes}} \\ \hline
			\textbf{EGO-Swarm} &
			39.070 &
			24.527 &
			1.124 &
			\multirow{3}{*}{\textbf{\underline{No}}} &
			\multirow{3}{*}{\textbf{\underline{Yes}}} &
			\multirow{3}{*}{\textbf{\underline{Yes}}} \\ \cline{1-4}
			\textbf{MINCO-Swarm} &
			25.955 &
			\textbf{\underline{24.102}} &
			0.673 &
			&
			&
			\\ \cline{1-4}
			\textbf{Proposed} &
			\textbf{\underline{24.124}} &
			24.111 &
			\textbf{\underline{0.427}} &
			&
			&
			\\ \hline
		\end{tabular}
	}
	 \vspace{-0.4cm}
\end{table*}

From Fig. \ref{pic:comparison_single}a, three optimization-based methods (Mapless, EGO-Planner, MINCO-Single) always show more pronounced twists and turns within a local area, while sampling-based methods (the proposed, MPL) generate smoother trajectories in trajectory shape. We think the reason is that the trajectory optimization problems formulated by the first three methods are always constrained by both the cluttered environment and nonlinear robot dynamics and are highly non-convex, thus the solution is always restricted within a local optimum, making the planner short-sighted. By contrast, sampling-based methods explore a larger freespace in candidate primitive evaluation. Compared to MPL, the proposed method generates a shorter trajectory own to the high quality of the motion primitive library. From Fig. \ref{pic:comparison_single}b, the proposed method managed to keep the maximum speed during the whole flight also thanks to the time-optimal primitives that have already handled the prior-known drone dynamics offline.

From the result, EGO-Planner tends to produce more conservative trajectories due to the convex hull constraint of B-spline. Mapless adopts a lightweight kd-tree data structure for environment representation, but the cost of rebuilding the kd-tree and the $O(\log n)$ time complexity of querying collision status cannot be neglected. Moreover, Mapless performs poorly in dense environments, as demonstrated in the attached video, due to its $O(n\log n)$ time complexity \cite{ji2021mapless} with respect to trajectory sampling, which limits its ability to find freespace. These methods always face a trade-off between environment representation and trajectory generation, both of whose computational cost cannot be simultaneously reduced in these methods.
In contrast, the proposed method eliminates the need for map inflation. Furthermore, the computation time is hardly affected by the number of primitives, as shown in Fig. \ref{pic:obs_collision_time} and \ref{pic:agent_collision_time}, thereby minimizing the online computational cost while ensuring high-quality trajectories.

\subsection{Swarm Simulation and Comparison}
\label{sec:swarm_comparison}

We compare the proposed method with RBP (2020)\cite{park2020efficient}, EGO-Swarm (2021)\cite{zhou2021ego}, MINCO-Swarm (2022)\cite{zhou2022swarm}, MADER (2021)\cite{tordesillas2021mader}, R-MADER (2023)\cite{kondo2023robust}, and AMSwarmX (2023)\cite{adajania2023amswarmx}. All of the above are open-source works that have demonstrated effectiveness in the field of aerial swarm planning in recent years.

Comparisons are conducted in both empty and cluttered space, to show the capability of inter-agent collision avoidance and avoid dense obstacles. The drone radius is set to 0.15 meters, the desired speed is set to 1 m/s, and all other parameters are maintained at their default values. For RBP, we adjust the \textit{plan\_time\_scale} parameter to achieve the best performance. Primitive number $N_{t}$ of the proposed method is set to $181$. The meanings of flight distance and computation time are consistent with those in Section \ref{sec:single_comparison}.

\begin{figure}[h]
	\centering
	\includegraphics[width=1.0\linewidth]{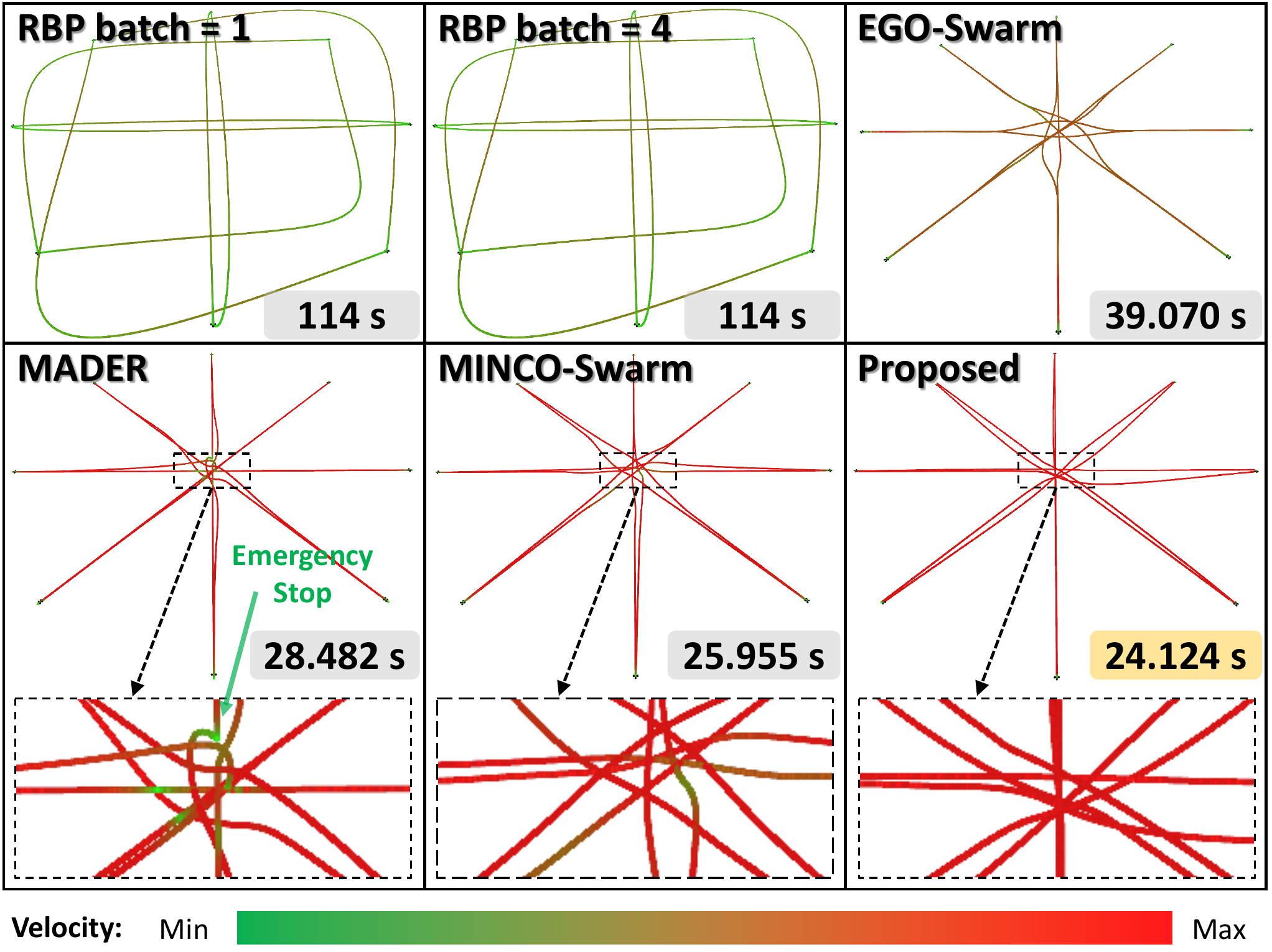}
    \caption{Trajectory comparisons of eight drones exchanging positions. The data in the bottom right corner shows the average flight time of the eight drones. The colored curves represent the trajectories of the drones. We use a color bar from green to red to show the change in drone speed from small to large.}
	\label{pic:bench_circle}
	\vspace{-0.4cm}
\end{figure}

\begin{figure}[h]
\centering
\includegraphics[width=1.0\linewidth]{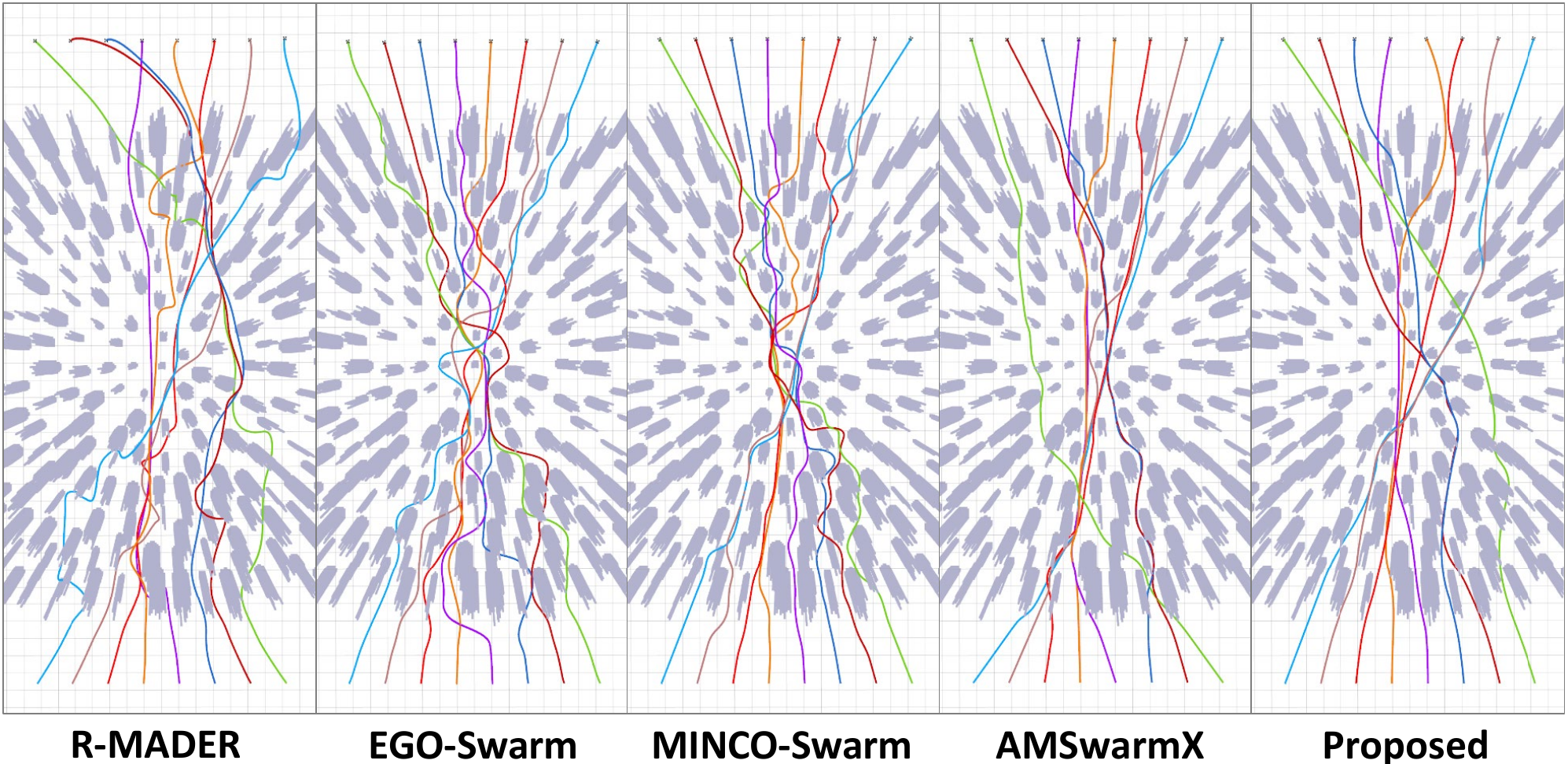}
\caption{Smoothness comparison by trajectory shape. Eight drones fly through the obstacle area from the bottom to the top of the image. All images are captured from a top-down perspective.}
\label{pic:traj_shape_bench}
\vspace{-0.6cm}
\end{figure}

\begin{figure*}[!h]
\centering
\includegraphics[width=1.0\linewidth]{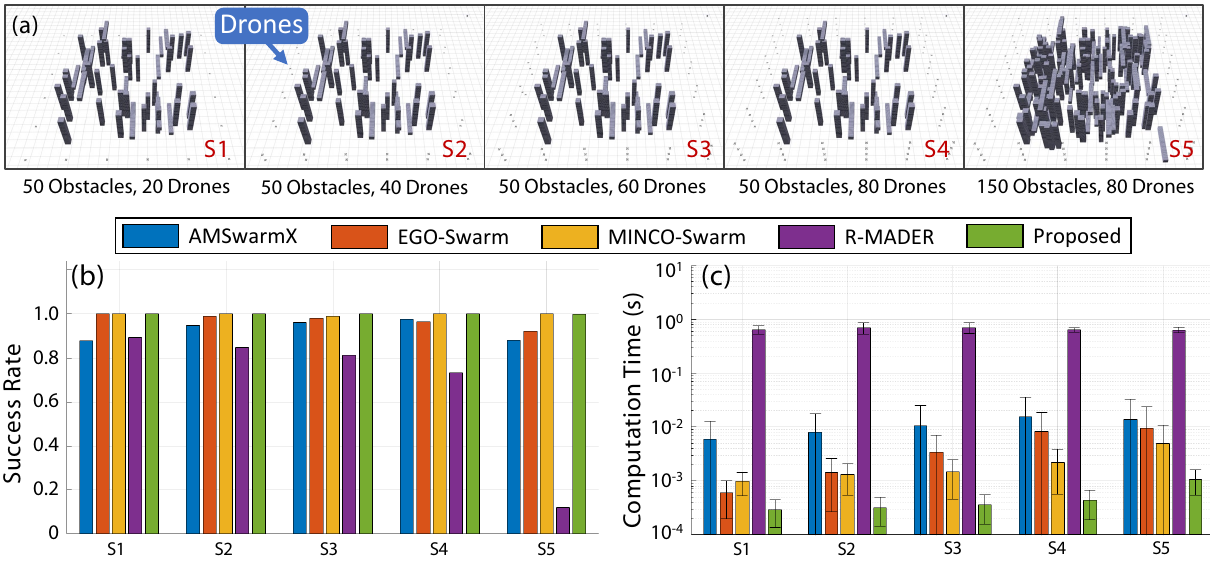}
\caption{Swarm planning comparison with varying drone number and obstacle number. Each test runs for 30 times and before each restart, the start and goal points are randomly reshuffled. (a) S1-S5: five experiment configurations. (b) Success rate comparison. (c) Computation time comparison.}
\label{pic:30_radnom_comp}
 \vspace{-0.6cm}
\end{figure*}

\subsubsection{Comparison without obstacles} Eight drones exchange positions in a circular empty space with a radius of $12~m$. The results are summarized in TABLE \ref{tab:comparison_empty}.
Since RBP is a centralized method, the computation time represents the total time of planning global trajectories for all drones, thus marked with *. For other methods, the computation time indicates the time of planning a local trajectory for each drone.
All data represent the average values across eight drones over $10$ experiments.
The executed trajectories of all methods are depicted in Fig. \ref{pic:bench_circle}. RBP and EGO-Swarm generate conservative trajectories due to the convex hull constraint of the Bernstein polynomial, leading to a compressed solution space.
MADER employs MINVO basis to alleviate this issue but still faces challenges in robot-robot avoidance, resulting in multiple stops when drones encounter each other, which can be seen in the attached video.
RBP incurs high computation cost, making it unsuitable for real-time planning on onboard devices.

\subsubsection{Comparison with obstacles} 
This part includes a qualitative experiment to evaluate the smoothness of the trajectory shape and a more in-depth quantitative experiment with multiple indicators. In this section, we choose EGO-Swarm (2021)\cite{zhou2021ego}, MINCO-Swarm (2022)\cite{zhou2022swarm}, R-MADER (2023)\cite{kondo2023robust}, and AMSwarmX (2023)\cite{adajania2023amswarmx} for comparison. RBP\cite{park2020efficient} is excluded because it is a centralized planner. MADER\cite{tordesillas2021mader} is also ignored because we have already chosen its successor R-MADER.

For the qualitative experiment, we tested these method with 200 cylindrical obstacles distributed in a $26\times20\times3 ~m$ space, as shown in Fig.~\ref{pic:traj_shape_bench}. The average radius of obstacles is $0.6~m$. Drone spacing of the start points and the spacing of the goals are both set to $2~m$, and the order of start points and goals is reversed.

From Fig.~\ref{pic:traj_shape_bench}, R-MADER, EGO-Swarm, MINCO-Swarm generated trajectories with a more tortuous shape. In this test, we modify the R-MADER parameter \textit{Ra (Radius of the planning sphere)} from $30 m$ to $7 m$, otherwise no solution can be found. A larger \textit{Ra} increases the dimension of the solution space, which is beneficial for finding solutions that are in line with long-term interests, but it also significantly increases the difficulty of optimization, especially when the number of constraints increases substantially in complex scenarios. $7~m$ is a balance point through continuous testing in this scenario. AMSwarmX and the proposed method both generated smoother trajectories. AMSwarmX is an alternating-minimization-based
approach whose key idea is pulling the trajectory to some attraction points $\mathbf{p}_{i,r}$ located in free space. To find the important $\mathbf{p}_{i,r}$ and for obstacle distance query, AMSwarmX uses A* search as the front end and requires a pre-built Octomap. This limits its application in unknown scenarios with onboard sensors. Furthermore, AMSwarmX is the only synchronous planner with information shared and trajectories planned at each same planning round\footnote{\url{https://github.com/utiasDSL/AMSwarmX/blob/master/amswarmx/src/runner/run_am_swarm.cpp}}. Part of the reason why the proposed method generates smoother trajectories is analysed in Sec. \ref{sec:single_comparison}.

Furthermore, we conducted more in-depth quantitative experiments. Several illustrative pictures are shown in Fig. \ref{pic:30_radnom_comp}.
The independent variables in this experiment are the number of drones and obstacles, the former is used to evaluate the ability of reciprocal collision avoidance, and the latter is for evaluating obstacles avoidance. The numbers of drones are set to 20, 40, 60, and 80, and the numbers of obstacles are set to 50 and 150. In the scenario with 150 obstacles, we only tested the largest 80-drone swarm to simultaneously evaluate the planner performance in extreme scenarios. \textit{Success Rate} and \textit{Computation Time} are the metrics we used. Among them, \textit{Success Rate} is defined as the ratio of drones that safely reach the goal point, \textit{Computation Time} is the time spent on a single trajectory planning. Each data point is the average value of all drones (20 to 80) in 30 random start-goal configurations. For \textit{Computation Time}, we used a logarithmic scale and plotted its standard deviation. R-MADER runs on the 80-core workstation, the same platform used for 1000-drone simulation in Sec. \ref{sec:large_scale}, since each agent of R-MADER requires about 1 CPU core. Other methods run on the 8-core-16-thread personal computer introduced in Sec. \ref{sec:details}.

A successful flight is defined as reaching the goal point without any collisions. For success rate comparison in Fig. \ref{pic:30_radnom_comp}b, all methods except R-MADER have shown a relatively high success rate. Among them, the success rate of EGO-Swarm gradually decreases with the increasing number of drones and obstacles, while that of AMSwarmX appears to be less related to the experimental setup. It is speculated that the main reason for this phenomenon is that AMSwarmX pre-builds the map and uses front-end global search to find high-quality feasible space for the back-end trajectory planner. The other methods (EGO-Swarm, MINCO-Swarm, The Proposed) only consider local information perceived by simulated sensors in real-time, so the success rate for later methods is easily affected by environment. In fact, the success rate of AMSwarmX is underestimated because, although AMSwarmX is a decentralized planner, in terms of code implementation, all drones run within the same program, and a collision of any drone will terminate the entire swarm. The success rate of R-MADER is heavily affected by the number of obstacles. A possible reason is that R-MADER uses a multitude of separation planes to separate free space from obstacles. This strategy leads to the issues of computational burden and conservatism with the increasing number of obstacles, which further results in long computation time and slower response to changing environment. Referring to the computation time depicted in Figure \ref{pic:30_radnom_comp}c, the proposed method exhibits a notable advantage, with computation times typically amounting to just one-third of those required by other methods. This result shows the effectiveness of the proposed method that transforms online trajectory optimization into a primitive selection problem.

\begin{figure}[h]
	\centering
	\includegraphics[width=1.0\linewidth]{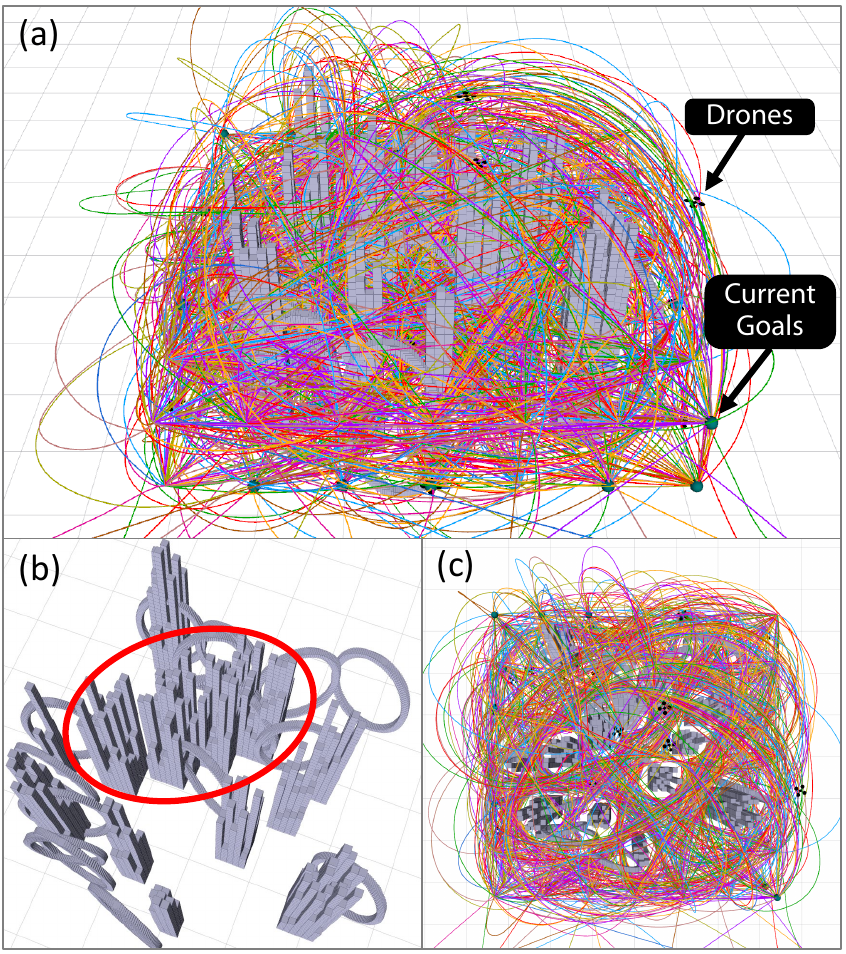}
	\caption{Long-time continuous flight. 20 drones fly in a $6\times6\times3~m$ cluttered space. Goals are randomly assigned once a drone reaches the current goal. Drones continuously fly for more than one hour in each round and repeat for 11 rounds with different obstacles. (a) and (c) represent trajectories 20 drones traveled over a period of approximately 10 minutes. The red circle in (b) shows a ``big wall" of randomly generated obstacles, making the collision avoidance more challenging.}
	\label{pic:20_drone}
	 \vspace{-0.5cm}
\end{figure}

\begin{figure}[t]
\centering
\includegraphics[width=1.0\linewidth]{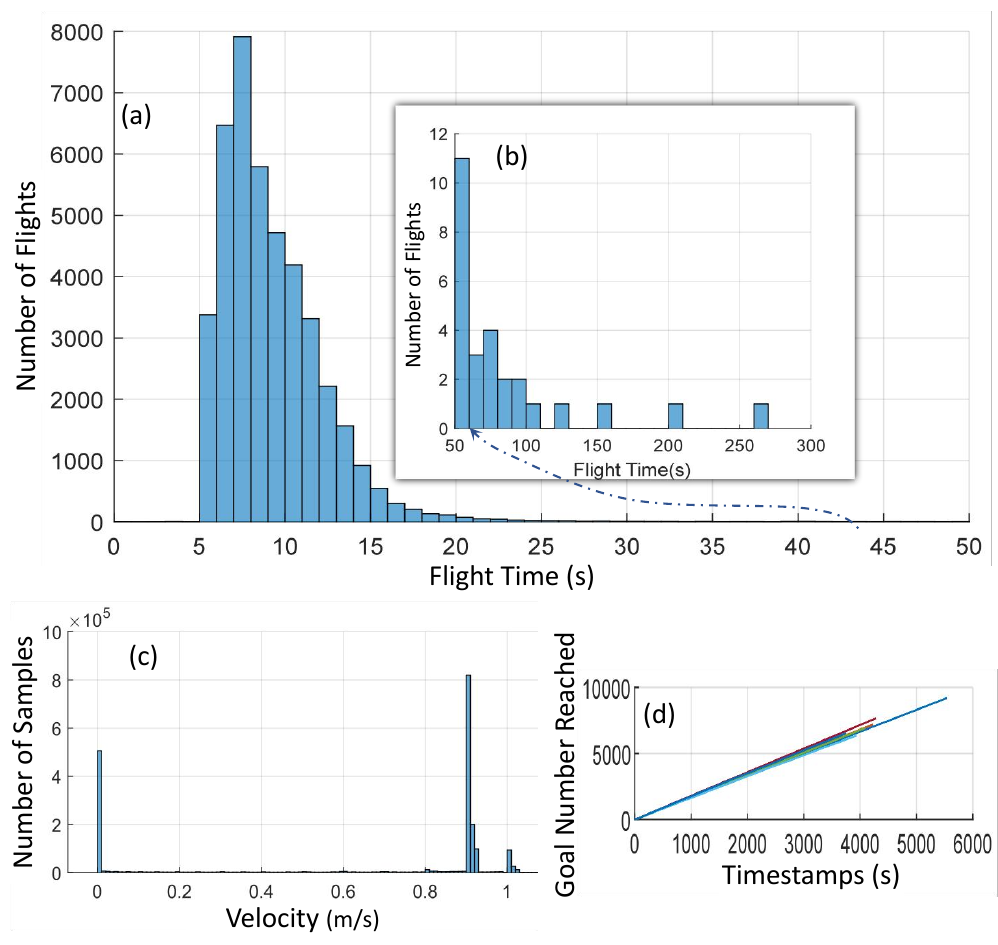}
\caption{(a) Histogram of the flight time used to reach a goal. (b) Cases of long flight time. Note that (a) and (b) are continuous on the horizontal axis. (c) Histogram of the flight velocity sampled at 10 Hz. (d) Goal number reached during the flight. Note that we omitted the legend, as the trends of all curves are essentially consistent.}
\label{pic:long_time}
 \vspace{-0.6cm}
\end{figure}

\subsection{Eleven Hours Continuous Flight}
\label{sec:11-hour}

This experiment is for evaluating the tolerance for occasional situations that are difficult to be modeled and specifically addressed. Due to the low probability of occurrence, long-term continuous flight testing is required. Existing swarm planning works have not yet conducted such tests, making it difficult to assess the robustness of their works against occasional events. In this experiment, we designed the following experiment: A dense obstacle scenario of $6\times6\times3~m$ is randomly generated along with 20 drones, and then a goal point setting software randomly sends goal points to each drone. Whenever a drone reaches a goal point, it will receive another random goal point. After at least 1 hour of continuous flight, the simulator randomly regenerates a new map and repeats the above process 11 times. This experiment also simulates an aerial logistics system, where the goal setting software simulates a logistics dispatch station, assigning delivery tasks to each drone. Each logistics aircraft needs to avoid obstacles (such as buildings or mountains) and other logistics aircraft during flight. 

The result is shown in Fig. \ref{pic:20_drone} and \ref{pic:long_time}, and the attached video \textit{11 Hours Flight}. We have compiled a histogram of the time taken to reach the goals. Flight time can effectively demonstrate the performance of the algorithm in long-duration navigation, as anomalies encountered during flight often lead to a significant increase in flight time. Moreover, for the delivery missions mentioned earlier, flight time is often one of the most critical metrics for users, provided that safety is ensured, which will be analyzed in Sec. \ref{sec:large_scale}. This section focuses more on the timeliness of the method. 

As shown in Fig. \ref{pic:long_time}, the vast majority of goal points can be reached within 20 seconds. For reference, the proportion of flights within 20 seconds is 99.1\%, and within 50 seconds is 99.94\%. For some flights with particularly long durations, readers can see at 15 seconds into the accompanying video \textit{11 Hours Flight}, in the lower left scene, a red trajectory and a yellow trajectory drone getting stuck in front of a big obstacle. Although they eventually escaped the blocking obstacle, a considerable amount of time was still spent. The root cause of this issue is that this paper, as a local planning algorithm, only considers the current environmental information and lacks the ability for long-term planning and decision-making. Therefore, it is more suitable for avoiding nearby moving or stationary obstacles, even if the obstacle number is very high. In contrast, AMSwarmX \cite{adajania2023amswarmx} in Sec. \ref{sec:swarm_comparison} introduces a global search based on A* to find the long-term path. The other works, including EGO-Swarm \cite{zhou2021ego}, MINCO-Swarm \cite{zhou2022swarm}, and R-MADER \cite{kondo2023robust}, are all local planners and are theoretically unsuitable for dealing with large obstacles or maze-like scenarios. Local planning algorithms, including the method in this paper, often need to be paired with a guidance front end responsible for high-level decision-making to form a complete autonomous navigation system for complex environments. In summary, the experiments in this section validate the robustness and reliability of the proposed method in cluttered environments, and can handle larger obstacles to some extent.

\begin{figure*}[!h]
	\centering
	\includegraphics[width=0.8\linewidth]{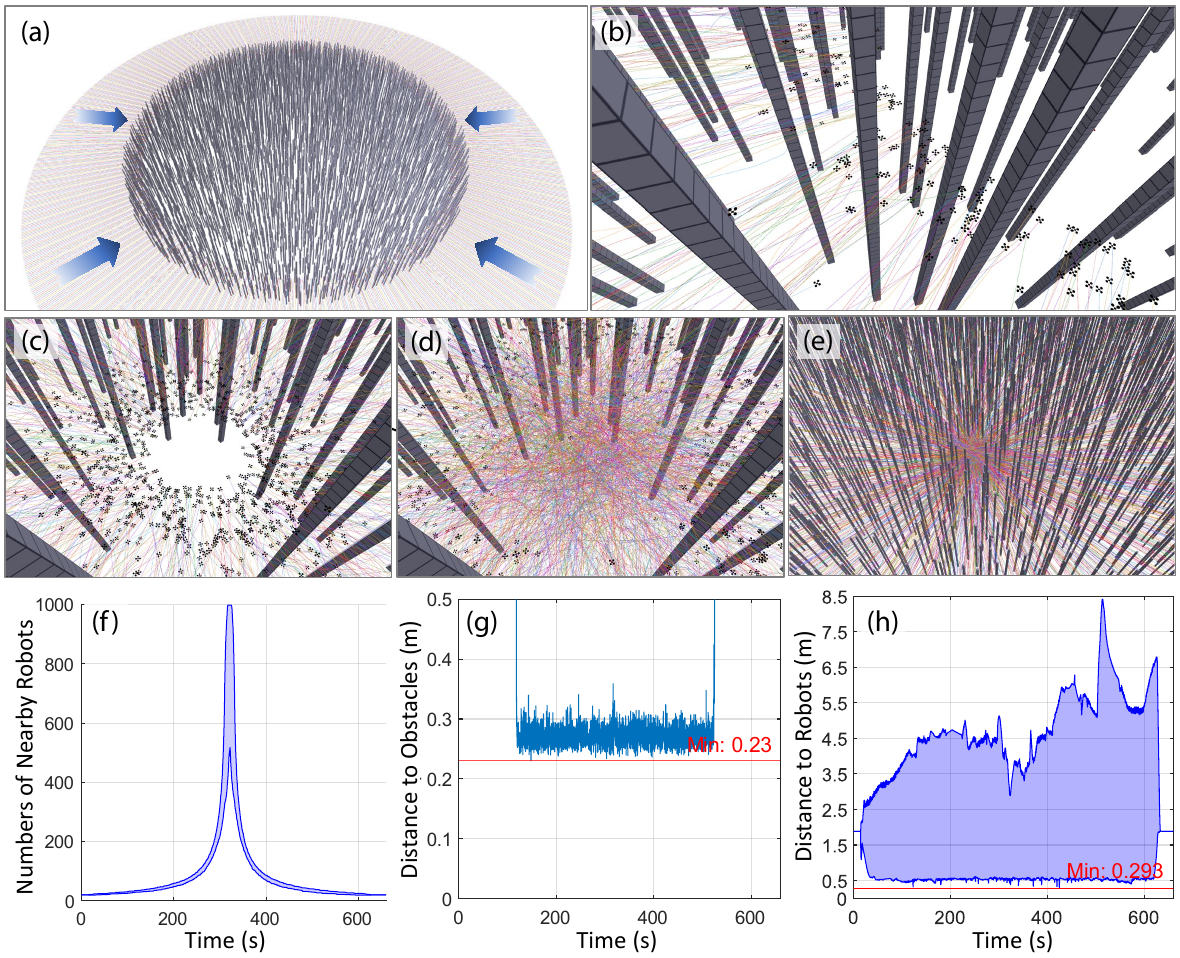}
	\caption{1000-drone position swap in cluttered environment. (a-e) The procedure of the flight. (f) The minimum-to-maximum number of nearby drones within a 20-meter radius around each drone. (g) The minimum distance to obstacles of all the drones at each timestamp. (h) The minimum-to-maximum drone-to-drone distance of each drone at every timestamp.}
	\label{pic:1000_circle}
	 \vspace{-0.4cm}
\end{figure*}

\subsection{The Large-scale Swarm Simulation}
\label{sec:large_scale}

Scalability is a critical metric for evaluating autonomous swarm algorithms. We demonstrate the capabilities of large-scale fully autonomous swarms consisting of $1000$ drones navigating in unknown environments.
Since each robot needs hundreds of megabytes of memory for simulated obstacle sensing and a lot computing for simulated drone motion in real-time, we conduct this experiment on a high-performance cloud workstation of model ecs.hfg6.20xlarge\footnote{\url{https://www.alibabacloud.com/solutions/sap?spm=a3c0i.23458820.2359477120.13.61d67d3fDISyrx}}. Its CPU contains 80 virtual cores running on Intel Xeon Platinum 8269CY and the memory size is 384 GB. Each robot runs in independent ROS (Robot Operation System) nodes to maximize the similarity to real-world deployment. In this evaluation, we replace the simulated depth map with point cloud acquired around each robot below $5~m$ distance, because depth rending consumes huge amount of GPU computing that only supercomputers can afford but will only make very slight differences to the results of trajectory planning. In other benchmark simulations and real-world experiments, the trajectories are shared via the loopback network and wireless network, respectively. However, in this large-scale simulation, we use inter-process memory sharing to share the trajectories as ROS can not support so many connections. The total CPU usage during the flight is about 80\%.

\begin{figure*}[!h]
	\centering
	\includegraphics[width=0.8\linewidth]{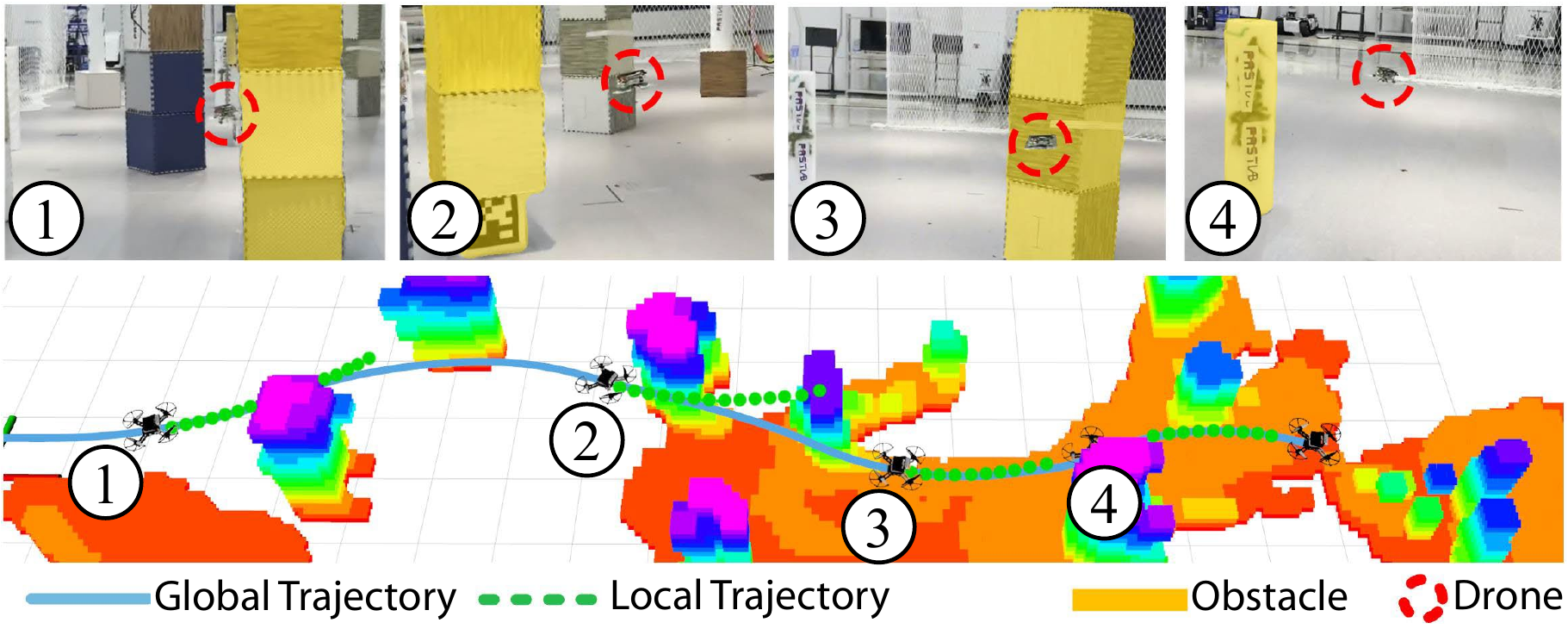}
	\vspace{0.1cm}
	\caption{Single-robot real-world experiment. The SWaP (Size, Weight, and Power) constrained quadrotor rapidly navigates through a series of obstacles one by one, achieving a maximum expected speed of $2m/s$ using time-optimal primitives.}
	\label{pic:single_real_world}
	\vspace{-0.5cm}
\end{figure*}

Fig. \ref{pic:1000_circle} illustrates the entire process of trajectory planning for $1000$ drones, which perform position exchange on a $300$-$m$-radius circle to maximize the collision avoidance. The initial spacing between adjacent drones is $1.88~m$. Velocity limit is set to $1~m/s$, primitive length $l$ is set to $6~m$, with $D_{angle} = 15^\circ$ and $N_a = 15$, which result in the number of primitives $N_{t} = 361$. The radius of the drone $r_{robot}$ is $0.15~m$. The corresponding video is \textit{The 1000-Drone Position Swap}.
Sub-figures (f) - (g) are the maximum or minimum values among all the drones at each timestamp. 
Fig. \ref{pic:1000_circle}f indicates that as all drones converge towards the center of the circle, the number of drones around each one gradually increases, reaching a peak of 1000 at the vicinity of the center around the $300~s$ timestamp. For the drones at the outermost edge, there are still at least 500 drones within a 20-meter radius. This situation implies the most challenging inter-robot collision avoidance. Even under such a dense distribution of obstacles and drones, the planner managed to keep a safe distance to obstacles and other robots, as shown in Fig. \ref{pic:1000_circle}g and h. The minimum distance in the sub-figure (h) is $0.293 m$, which is slightly shorter than the $0.3~m$ inter-robot safe clearance defined as twice the radius of the drone. This indicates that the current drone density is nearly reaching the upper limit of the planner's capabilities. In sub-figure (h), the maximum inter-robot distance reduces to about $2.5~m$ near the timestamp of $300 s$. In fact, for a global optimal solution, both the maximum and minimum distance should be close to the $0.3~m$ spacing constraint. This implies that all drones achieve the densest arrangement, which also means the shortest flight distance. In contrast, the proposed method in this paper, being a real-time planning algorithm, only has local environmental information within a few seconds into the future, hence the generated trajectories are relatively conservative.

In Fig. \ref{pic:E1000}, we conduct another test where drones are separated into 5 layers. The start and the goal points are both located on the circle with a radius of 300 meters. The quasi-linear appearance of accumulated trajectories stems from the vast scale disparity between the experimental area and drones. Actually, each drone dynamically selects collision-free trajectories from our motion primitive library in real-time. The total flight processes can be observed in the attached video. The results demonstrate that our proposed method excels in swarm scalability, showcasing its potential for handling large-scale swarm scenarios effectively.

\begin{figure}[t]
	\centering
	\includegraphics[width=1.0\linewidth]{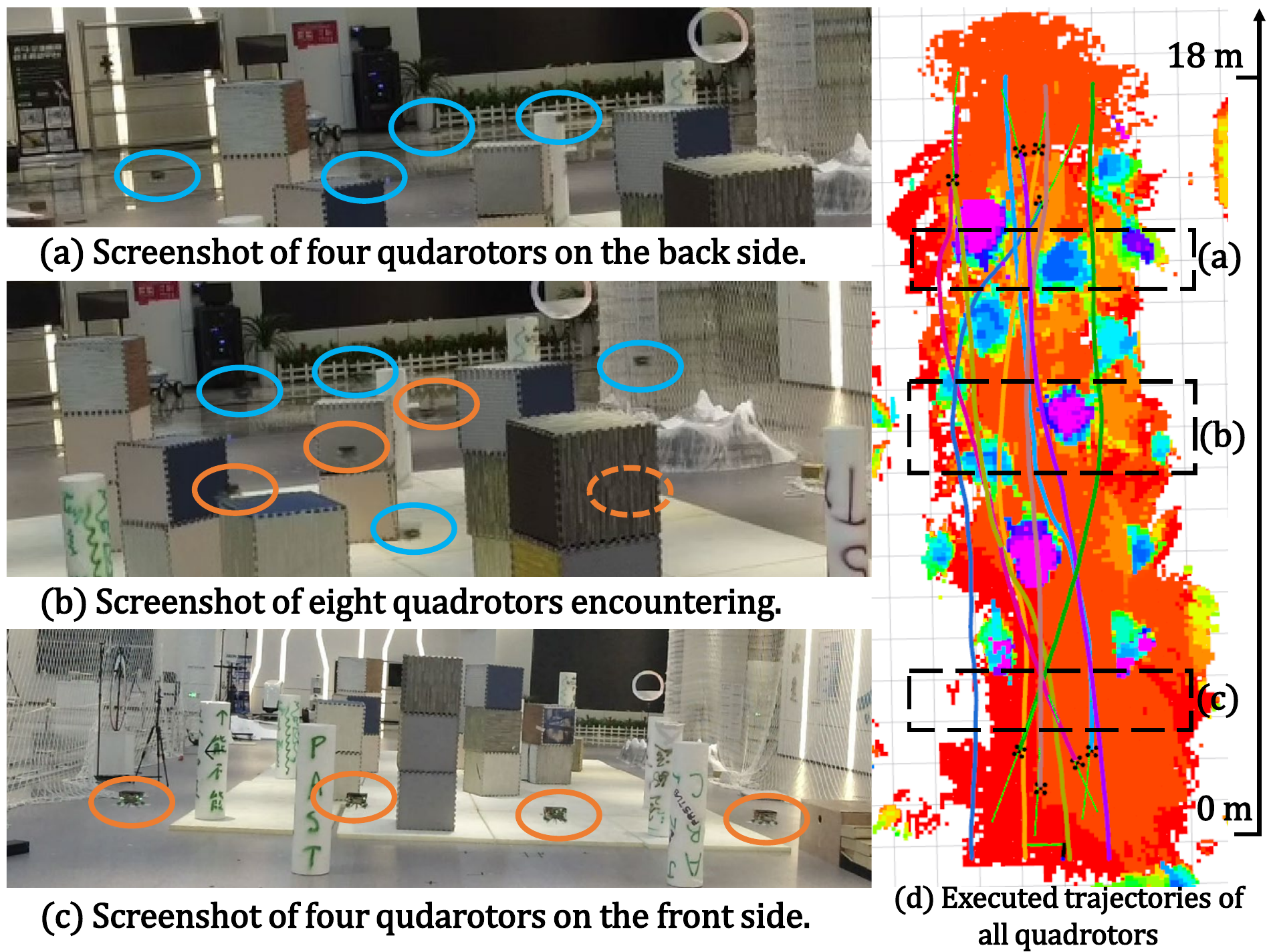}
	\caption{Swarm real-world experiment. Colored ellipses indicate the positions of quadcopters. The orange ellipses mark the four drones that fly forward and the blue ellipses mark the four drones that fly backward. The orange dashed ellipse represents a quadcopter behind an obstacle. The dashed box indicates the screenshot location. Colored curves depict the executed trajectories.}
	\label{pic:FII}
	\vspace{-0.4cm}
\end{figure}

For optimization-based methods that necessitates providing a well-defined initial guess for nonlinear trajectory optimization in non-convex freespace, as the number of drones increases along with their trajectories, the environment becomes partitioned into numerous sub-spaces, each with unique local minimum. This division poses challenges in trajectory optimization, especially in determining a suitable initial guess, making it difficult to find a suitable local minimum or even just a safe solution. Consequently, the scalability of swarms becomes constrained.
In contrast, the proposed method operates directly on the raw point cloud without the need for additional environment representation and obstacle inflation. Moreover, collision checking and the linear-complexity trajectory selection are completely decoupled, ensuring that they do not affect each other. This allows for the use of a big motion primitive library to cover the entire reachable space simultaneously, greatly enhancing the scalability of the swarm. As a result, our swarm algorithm is ultra-lightweight and well-suited for deployment in large-scale swarms.

\subsection{Real-world Experiments}
\label{sec:real_world}

We conduct real-world experiments to validate the proposed method in both single-robot and swarm scenarios. The planning algorithm is deployed on SWaP (Size, Weight, and Power) constrained quadrotor platforms.

An offline motion primitive library is generated, containing $109$ time-optimal and dynamically feasible trajectories. The library parameters include $N_a = 7$ arcs, different radii $r \in\{2,3,4,6,8,12,20,36,78,\infty\}m$, length $l = 3m$, 
different start angles $\theta \in \{0^\circ, -10^\circ, -20^\circ, 0^\circ, -10^\circ,$ $ -20^\circ, 0^\circ, -10^\circ, -20^\circ, 0^\circ\}$, and rotation angle interpolation $D_{angle} = 30^\circ$. The environments consist of various obstacles randomly placed, including cubes and cylinders, with an average distance of around $1~m$ between them, creating challenging navigation scenarios for the quadrotors, especially in swarm experiments.
Importantly, all experiments are conducted without using any external localization or computing devices. All state estimation, planning, control, and communication modules run solely on the onboard computer. Each quadrotor independently and asynchronously executes planning tasks in unknown environments, maintaining a decentralized autonomous system architecture.

\subsubsection{Single-robot Experiments}

We perform two single-robot autonomous navigation experiments with quadrotors, setting the speed and acceleration limits to $2m/s$ and $6m/s^2$, respectively.
In the indoor experiment, the quadrotor's global goal is set $18~m$ ahead. It rapidly avoids obstacles one by one at a speed of almost $2m/s$. Fig. \ref{pic:single_real_world} displays screenshots capturing the entire flight process.
In the outdoor experiment, the quadrotor's global goal is set $40~m$ ahead. It successfully avoids multiple trees at an approximate speed of $2m/s$ in an unknown natural environment. The results is in the attached video.
These two experiments demonstrate the practicality of the proposed method in single-robot autonomous navigation.

\subsubsection{Swarm Experiments}

In the swarm real-world navigation experiments, we set the maximum speed and acceleration of each quadrotor to $1m/s$ and $3m/s^2$, respectively.
We conduct a bidirectional cross-flight experiment as depicted in Fig. \ref{pic:FII}. 
Four drones fly towards the goal of the 18m in front and four drones fly back, they will cross in the middle of the scenario.
The screenshots of their start and encountering positions are shown in Fig. \ref{pic:FII}(a)-(c), and the trajectories of all quadrotors are presented in Fig. \ref{pic:FII}(d). 
Fig. \ref{pic:FII}(a)-(c) illustrate screenshots of their starting and encountering positions, and Fig. \ref{pic:FII}(d) presents the trajectories of all quadrotors. They fly towards their respective goals at an approximate speed of $1m/s$. The experiment confirms the practicality of the proposed planner in swarm scenarios.

\section{Discussion and Conclusion}
\label{sec:conclusion}

For time and space consumption in simulations and experiments, the offline generation process of the motion primitive library takes about $38.6s$ and outputs a $75.3\rm{MB}$ motion primitive library when $N_t = 109$. This library size is easily accommodated in modern RAM (random-access memory) with several gigabytes of storage. 
In practice, only the occupancy relationship $\mathcal{R}_o$, $\mathcal{R}_t$, and other essential information like the end position $\mathbf{p}_{end}$ need to be loaded into RAM for fast access. The majority of the library that describes the exact shape of each trajectory can be stored in low-speed and cheap storage such as an SD card.

In multi-agent literature, movement oscillation caused by deadlock is a common issue \cite{decastro2018collision, van2011reciprocal}, where agents generate commands for each robot itself without having a high-level coordination. Fortunately however, this issue is naturally mitigated by our method, owing to real-world randomness, asynchronously triggered planning, and a wide space coverage of our trajectory representation. Similarly, deadlock is also reported to hardly occur in other trajectory planning methods \cite{alonso2015collision, zhou2022swarm} in normal situations.

Acceleration discontinuity is a flaw of this work, which arises because replanning can be triggered at any time, and the acceleration at that moment cannot be anticipated when generating motion primitives. A feasible solution to achieve continuity is to disallow triggering replanning at any time and only allow switching to the next primitive after the execution of the current primitive is completed, while also fixing the derivatives of the initial and final states of all primitives to be some predefined common values. However, the cost of this strategy is a slower response of the planner to dynamic environments. 

To conclude this paper, we present an ultra-lightweight and scalable planner designed for large-scale autonomous aerial swarms. The planner addresses the high-dimensional trajectory generation problem by transforming it into several linear-complexity selection problems. By doing so, it minimizes online computational costs and is capable of being deployed on large-scale swarms, accommodating up to 1000 robots.

%\addtolength{\textheight}{-15cm}   

\bibliography{TRO_HJL}

\vspace{-0.2cm}
% 侯嘉良
\begin{IEEEbiography}[{\includegraphics[width=1in,height=1.25in,clip,keepaspectratio]{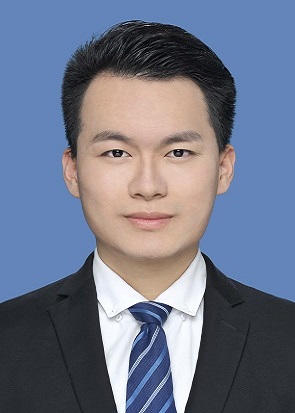}}]{Jialiang Hou} 
	received the B.Eng. degree in process equipment and control engineering from East China University of Science and Technology, Shanghai, China, in 2019. He is currently working toward the Ph.D. degree in computer application technology with Fudan University, Shanghai, China. His research interests include motion planning for aerial swarm robots and autonomous navigation.
\end{IEEEbiography}

\vspace{-0.2cm}
% 周鑫
\begin{IEEEbiography}[{\includegraphics[width=1in,height=1.25in,clip,keepaspectratio]{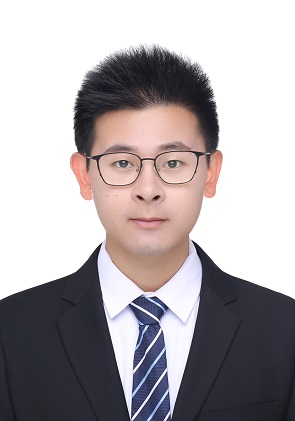}}]{Xin Zhou}
	received the B.Eng. degree in electrical engineering and automation from China University of Mining and Technology, Xuzhou, China, in 2019. He obtained his Ph.D. degree in electronic information from Zhejiang University, Hangzhou, China, in 2024. He is currently a postdoctoral fellow at the Hong Kong University of Science and Technology. His research interests include motion planning and mapping for aerial swarm robotics.
\end{IEEEbiography}

\vspace{-0.2cm}
% 潘能
\begin{IEEEbiography}[{\includegraphics[width=1in,height=1.25in,clip,keepaspectratio]{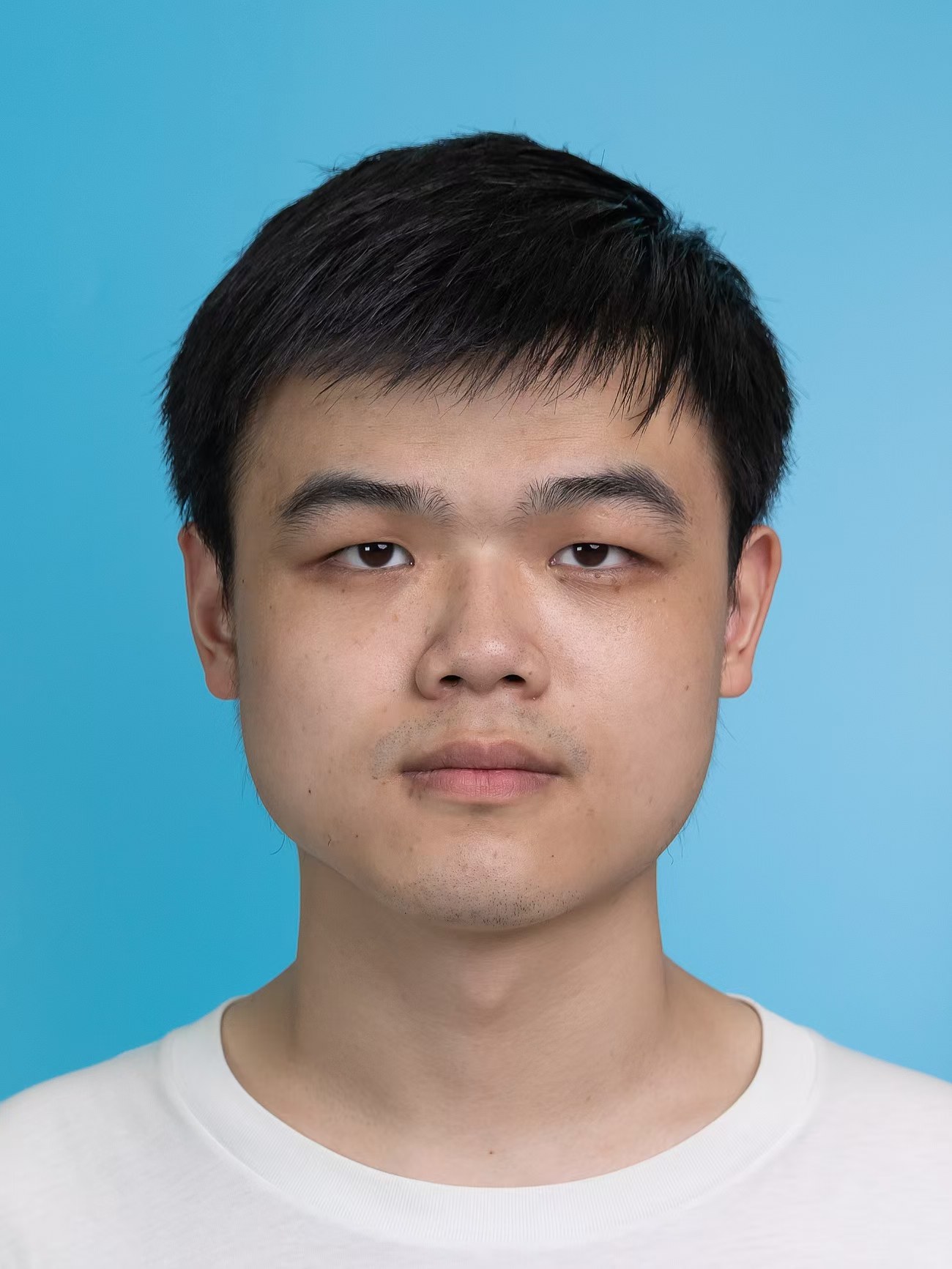}}]{Neng Pan} 
	received the B.Eng degree in control engineering from Zhejiang University, Hangzhou, China, in 2021, and the M.Eng degree in electronic information engineering from Zhejiang University, Hangzhou, China, in 2024 .He is currently a robotic engineer at Skysys Technology Co., Ltd. His research interests include uav design, control and planning.
\end{IEEEbiography}

\vspace{-0.2cm}
% 李昂
\begin{IEEEbiography}[{\includegraphics[width=1in,height=1.25in,clip,keepaspectratio]{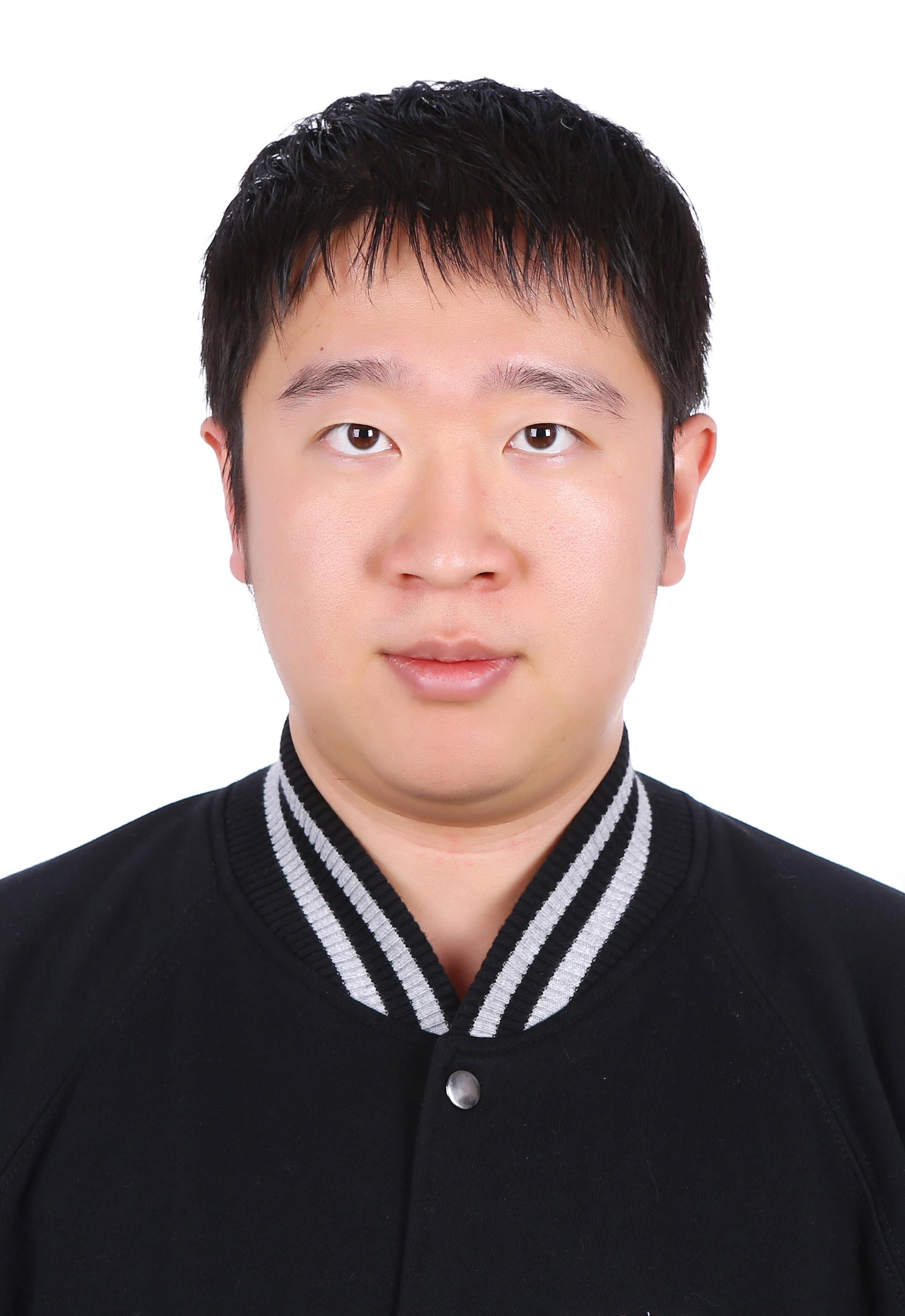}}]{Ang Li}
	received the B.S. degree in Aircraft Design and Engineering from Northwestern Polytechnical University, Xian, China, in 2018. He is currently working toward the Ph.D. degree in Aircraft Design at Beihang University, Beijing, China. His current interests include motion planning, aerial swarm, and target tracking.
\end{IEEEbiography}

\vspace{-0.2cm}
% 管宇翔
\begin{IEEEbiography}[{\includegraphics[width=1in,height=1.25in,clip,keepaspectratio]{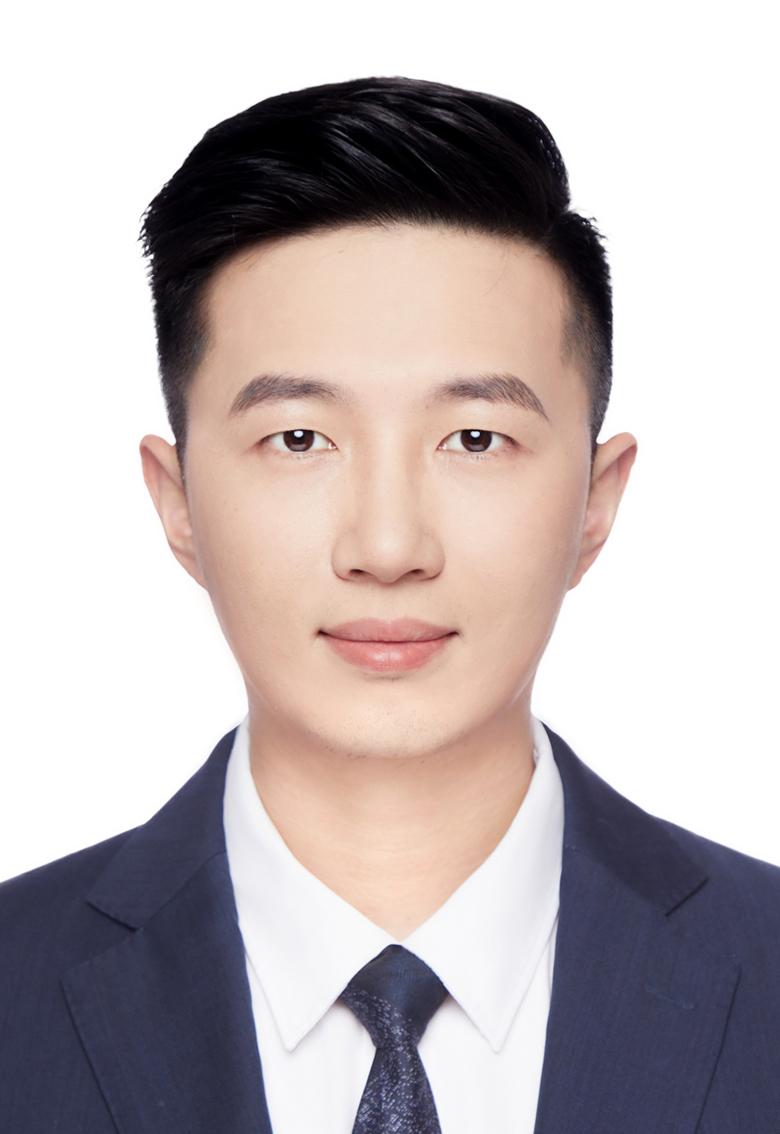}}]{Yuxiang Guan}
	received B.S. degree and M.Sc. degree in Environment Engineering from Keele University, Staffordshire, UK, in 2014 and 2015 respectively. He received Ph. D degree in Electronic Information Engineering from Fudan University, Shanghai, China, in 2025. His current research interests include multi-agent dynamic and task scheduling and allocation.
\end{IEEEbiography}

\vspace{-0.2cm}
% 许超
\begin{IEEEbiography}[{\includegraphics[width=1in,height=1.25in,clip,keepaspectratio]{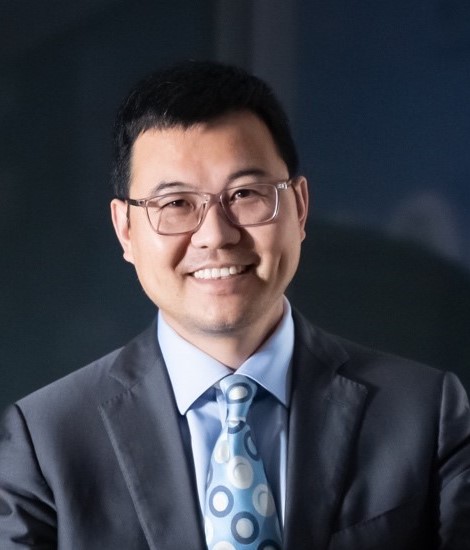}}]{Chao Xu}
	received the Ph.D. degree in Mechanical Engineering from Lehigh University in 2010. He is currently Associate Dean and Professor at the College of Control Science and Engineering, Zhejiang University. He serves as the inaugural Dean of ZJU Huzhou Institute, as well as plays the role of the Managing Editor for \textit{IET Cyber-Systems \& Robotics}. 
	His research expertise is Flying Robotics, Control-theoretic Learning. Prof. Xu has published over 100 papers in international journals, including \textit{Science Robotics}, \textit{Nature Machine Intelligence}, etc.
\end{IEEEbiography}

\vspace{-0.2cm}
% 甘中学
\begin{IEEEbiography}[{\includegraphics[width=1in,height=1.25in,clip,keepaspectratio]{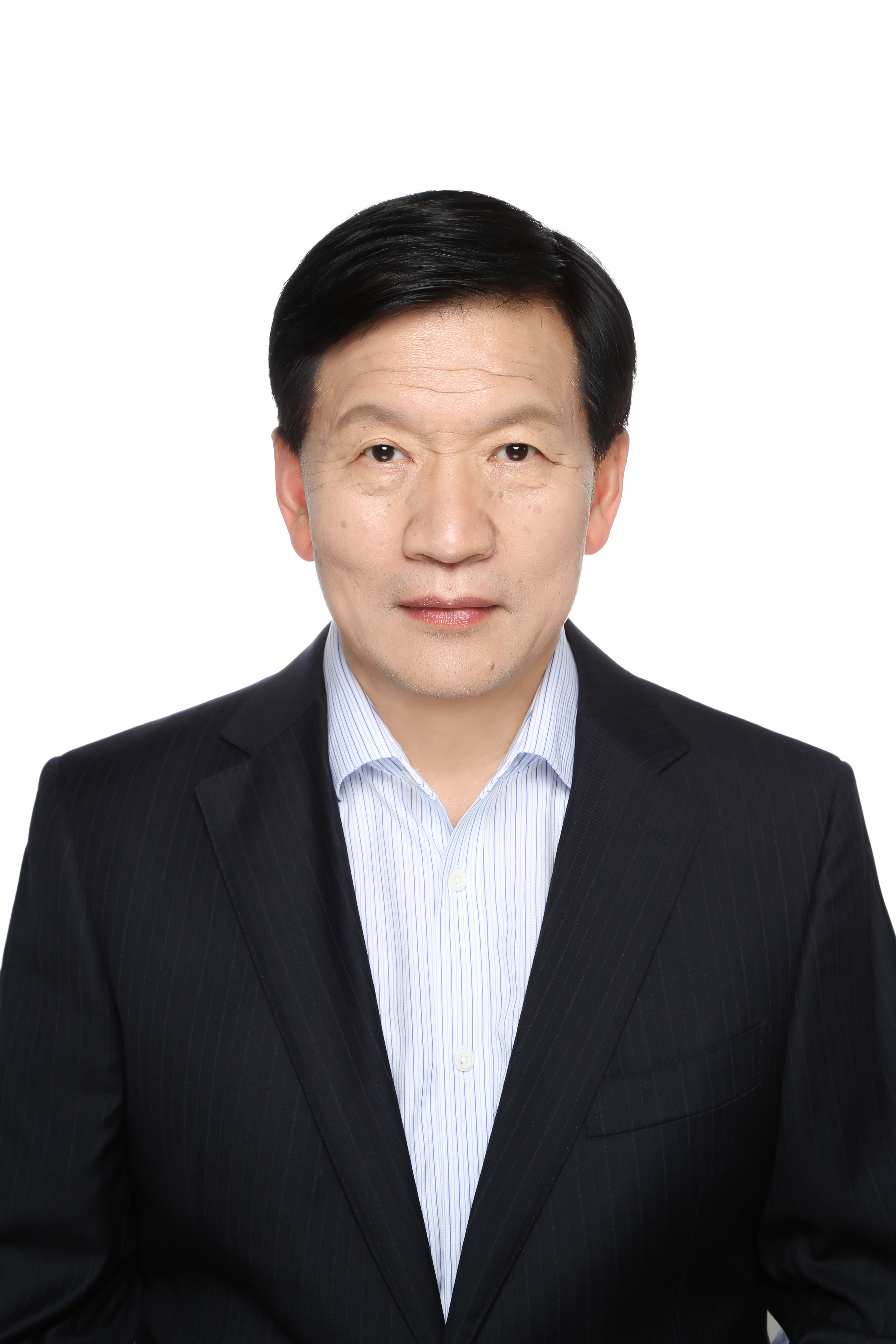}}]{Zhongxue Gan}
	received Ph.D. degree in mechanical engineering from University of Connecticut, Connecticut, USA, in 1993. He was a research fellow of ABB company during 1990 to 2005, and he was promoted to Chief scientist of ABB Global Robotics and flexible Automation in 2002. From 2006 to 2014, He was a Chief scientist and Vice Chairman of the Board at ENN Group, Langfang, China. Since 2017, he has served as Distinguished Professor in Fudan University, and been the dean of Institute of AI and Robotics, Fudan University. In 2010, he was awarded 'China International Science and Technology Cooperation Award' by Ministry of Science and Technology of the People's Republic of China.
\end{IEEEbiography}

\vspace{-0.2cm}
% 高飞
\begin{IEEEbiography}[{\includegraphics[width=1in,height=1.25in,clip,keepaspectratio]{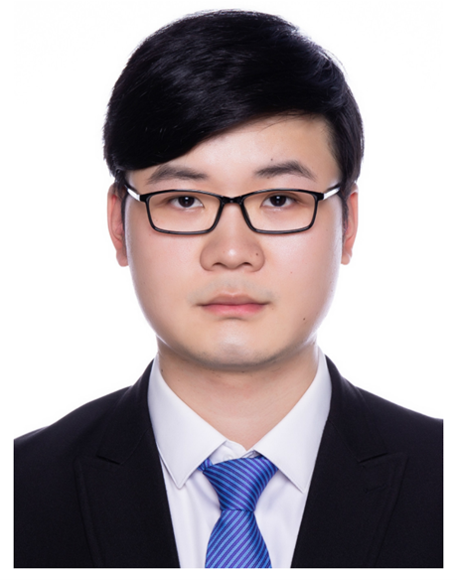}}]{Fei Gao}
	received the Ph.D. degree in electronic and computer engineering from the Hong Kong University of Science and Technology, Hong Kong, in 2019.  
	He is currently a tenured associate professor at the Department of Control Science and Engineering, Zhejiang University, where he leads the Flying Autonomous Robotics (FAR) group affiliated with the Field Autonomous System and Computing (FAST) Laboratory. 
	His research interests include aerial robots, autonomous navigation, motion planning, optimization, and localization and mapping. 
\end{IEEEbiography}

\end{document}